\documentclass[12pt, a4paper]{article} 
\usepackage{amsfonts,amsmath,amsthm,graphicx,float,fancyhdr,multirow,hyperref,color,booktabs,longtable,authblk}
\usepackage{mathrsfs,hhline}
\newtheorem{lemma}{Lemma}
\newtheorem{corollary}{Corollary}
\newtheorem{proposition}{Proposition}
\newtheorem{theorem}{Theorem}
\newtheorem{remark}{Remark}
\newtheorem{definition}{Definition}
\title{Towards a Mathematical Foundation of Immunology and Amino Acid Chains\thanks{The work
described in this paper is supported by GRF grant [Project No. 9041544] and [Project No. CityU 103210]
and [Project No. 9380050]. Email addresses: {\tt wenjshen@student.cityu.edu.hk} (W.-J. Shen), 
{\tt cshswong@cityu.edu.hk} (H.-S. Wong), {\tt qwxiao@live.com} (Q.-W. Xiao), 
{\tt xinguo2@cityu.edu.hk} (X. Guo), and {\tt smale@cityu.edu.hk} (S. Smale)}}
\author[1]{Wen-Jun Shen}
\author[1]{Hau-San Wong}
\author[2]{Quan-Wu Xiao}
\author[2]{Xin Guo}
\author[2]{Stephen Smale}
\affil[1]{Department of Computer Sciences, City University of Hong Kong}
\affil[2]{Department of Mathematics, City University of Hong Kong}

\begin{document}

\maketitle

\begin{abstract}
We attempt to set a mathematical foundation of immunology and amino acid
chains. To measure the similarities of these chains, a kernel on strings is defined
using only the sequence of the chains and a good amino acid substitution matrix 
(e.g. BLOSUM62). The kernel is used in learning machines to predict binding 
affinities of peptides to human leukocyte antigen DR (HLA-DR) molecules.
On both fixed allele \cite{Nielsen2009}
and pan-allele \cite{Nielsen2010} benchmark databases, our algorithm achieves the 
state-of-the-art performance. The kernel is also used to define a distance on an HLA-DR allele set
based on which a clustering analysis precisely recovers the serotype 
classifications assigned by WHO \cite{Holdsworth2009, Marsh2010}. These results suggest that our 
kernel relates well the chain structure of both peptides and 
HLA-DR molecules to their biological functions, and that it offers a simple, 
powerful and promising methodology to immunology and amino acid chain studies.
\end{abstract}

\section{Introduction}\label{SectionIntroduction}

Large scientific and industrial enterprises are engaged in efforts to
produce new vaccines from synthetic peptides. The study of peptide binding
to appropriate alleles is a major part of this effort. Our goal here is to
support the use of a certain ``string kernel'' for peptide binding prediction
as well as for the classification of supertypes of the major histocompatibility complex ({\bf MHC}) alleles.

Our point of view is that some key biological
information is contained in just two places: first, in a similarity kernel
(or substitution matrix) on the set of the fundamental amino acids; and
second, on a good representation of the relevant alleles as strings of these
amino acids. Our results bear this out.

This is achieved with great simplicity and predictive power. Along the way
we find that gaps and their penalties in the string kernels don't help, and
that emphasizing peptide binding as a real-valued function rather than a
binding/non-binding dichotomy clarifies the issues. We use a modification
of BLOSUM62 followed by a Hadamard power. We also use regularized least
squares ({\bf RLS}) in contrast to support vector machines as the 
former is consistent with our regression emphasis.

We next briefly describe the construction (more details also in Section \ref{section2})
of our main kernel $\hat K^3$ on amino acid chains,
inspired by local alignment kernels (see e.g. \cite{Saigo2006}) as well as an
analogous kernel in vision (see \cite{Smale2010}) begins.

For the purposes of this paper, a kernel $K$ is a symmetric function $K: X \times X \to 
\mathbb R$ where $X$ is a finite set. Given an order on $X$, $K$ may be 
represented as a matrix (think of $X$ as the set of indices of the matrix 
elements). Then it is assumed that $K$ is positive definite (in such a representation).

Let $\mathscr A$ be the set of the 20 basic (for life) amino acids.
Every protein has a representation as a string of elements of $\mathscr A$.

\begin{description}
 \item[Step 1. ] Definition of a kernel $K^1: \mathscr{A} \times 
 \mathscr{A} \to \mathbb{R}$. 

BLOSUM62 is a similarity (or substitution) matrix on $\mathscr A$ frequently used in 
immunology \cite{Henikoff1992}. In the formulation of BLOSUM62, a kernel $Q:\mathscr A 
\times \mathscr A \to \mathbb R$
is defined using blocks of aligned strings of amino acids representing 
proteins. One can think $Q$ 
as the ``raw data'' of BLOSUM62.
It is symmetric, positive-valued, and a 
probability measure on $\mathscr A \times \mathscr A$. (We have in addition checked
that it is positive definite.)

Let $p$ be the marginal probability defined on $\mathscr A$ by $Q$. That is,
$$
p(x) = \sum_{y \in \mathscr A} Q(x,y).
$$

Next, we define the {\bf BLOSUM62-2} matrix, indexed by the set $\mathscr{A}$, as
\begin{eqnarray*}
[\mbox{BLOSUM62-2}](x,y) = \frac{ Q(x,y) }{ p(x)p(y) }.
\end{eqnarray*}
We list the BLOSUM62-2 matrix in Appendix \ref{blosumDataAppendix}. 
Suppose $\beta >0$ is a parameter, usually chosen about 
$\frac{1}{8}$ or $\frac{1}{10}$ (still mysterious). Then a 
kernel $K^1: \mathscr{A} \times \mathscr{A} \to \mathbb{R}$ is given by
\begin{eqnarray}\label{local-KPower}
K^1(x,y) = \left( [\mbox{BLOSUM62-2}](x,y) \right)^\beta.
\end{eqnarray}
Note that the power in (\ref{local-KPower}) is of the matrix entries, not of the matrix.

 \item[Step 2.] Let $\mathscr A^1 = \mathscr A$ and define 
 $\mathscr A^{k+1} = \mathscr{A}^k \times \mathscr A$ recursively 
 for any $k \in \mathbb N$. We say $s$ is an amino acid chain (or string) if $s \in 
 \cup_{k=1}^\infty \mathscr A^k$, and $s = (s_1, \ldots, s_k)$ 
 is a $k$-mer if $s \in \mathscr{A}^k$ for some $k \in \mathbb N$ 
 with $s_i \in \mathscr A$. Consider
$$
K^2_k(u,v) = \prod^k_{i=1} K^1(u_i,v_i)
$$
where $u,v$ are amino acid strings of the same length $k$, $u=(u_1,
\ldots, u_k)$, $v=(v_1, \ldots, v_k)$; $u,v$ are $k$-mers.
$K^2_k$ is a kernel on the set of all $k$-mers.

\item[Step 3.] Let $f=(f_1,\cdots,f_m)$ be an amino acid chain. Denote by
$|f|$ the length of $f$ (so here $|f|=m$).
Write $u\subset f$ whenever $u$ is of the form
$u=(f_{i+1}, \cdots, f_{i+k})$ for some $1\leq i+1\leq i+k\leq m$.
Let $g$ be another amino acid chain, then define
$$
K^3(f,g) = \sum_{\substack{u \subset f, v\subset g \\ |u|=|v|=k \\ 
\text{all } k=1,2,\ldots}} K_k^2(u,v),
$$
for $f$ and $g$ in any finite set $X$ of amino acid chains.
Here, and in all of this paper, we abuse the notation to let 
the sum count each occurrence of $u$ in $f$ (and of $v$ in $g$).
In other words we count these occurrences ``with multiplicity''.
While $u$ and $v$ need to have the same length, not so for $f$ and $g$.
Replacing the sum by an average gives a different but related kernel.

We define the correlation kernel $\hat K$ normalized from any kernel $K$ by
$$
\hat K(x, y) = \frac{K(x,y)}{\sqrt{K(x,x) K(y,y)}}.
$$
In particular, let $\hat K^3$ be the correlation kernel of $K^3$.
\end{description}

\begin{remark}
$\hat K^3$ is a kernel (see Section \ref{KernelOnStrings}). It is symmetric, positive definite, 
positive-valued; it is basic for the results and development of this paper.
We sometimes say string kernel. The construction works for any kernel (at the place of  $K^1$)
on any finite alphabet (replacing $\mathscr{A}$).
\end{remark}

\begin{remark}
For some background see \cite{Haussler1999, Saigo2004,
Salomon2006, Leslie2002}. But we use no gap 
penalty or even gaps, no logarithms, no implied round-offs, and no alignments (except the 
BLOSUM62-2 matrix which indirectly contains some alignment information). 
Our numerical experiments indicate that these don't help in our context, (at least!).
\end{remark}

\begin{remark}
For complexity reasons one may limit the values of $k$ in 
Step 3 with a small loss of accuracy, or even choose the $k$-mers at random.
\end{remark}

\begin{remark}
The chains we use are proteins, peptides, and alleles.
Peptides are short chain fragments of proteins.
Alleles are realizations of genes in living organisms
varying with the individual; as proteins they have representations as amino acid chains.
\end{remark}

MHC II and MHC I are sets of alleles which are associated with immunological 
responses to viruses, bacteria, peptides and related. See \cite{Lund2005, Graur2000} for 
good introductions. In this paper we only study HLA II, the MHC II in human beings.
HLA-DRB (or simply {\bf DRB}) describes a subset of HLA II alleles which play a central role in immunology,
as well as in this paper.

\subsection{First Application: Binding Affinity Prediction}

Peptide binding to a fixed HLA II (and HLA I as well) molecule (or an allele) $a$ is a crucial
step in the immune response of the human body to a pathogen or a peptide-based
vaccine. Its prediction is computed from data of the form $(x_i,
y_i)_{i=1}^{m}$, $x_i\in \mathscr{P}_a$ and $y_i\in [0, 1]$, where $\mathscr{P}_a$ is a
set of peptides (i.e. chains of amino acids; in this paper we study 
peptides of length 9 to 37 amino acids, usually about 15) associated to an HLA II allele $a$. 
Here $y_i$ expresses the strength of the binding of $x_i$ to $a$.
The peptide binding
problem occupies much research. We may use our kernel $\hat K^3$
described above for this problem since peptides are represented as
strings of amino acids. Our prediction thus uses only the amino acid chains
of the peptides, a substitution matrix, and some existing binding affinities (as ``data'').

Following RLS supervised learning with kernel $K = \hat{K}^3$,
the main construction is to compute 
\begin{equation} \label{rls}
f_a = \arg \min_{f\in
\mathscr{H}_{ K }} \sum_{i=1}^{m}(f(x_i)-y_i)^2+\lambda\|f\|_{ K }^2 .
\end{equation}
Here $\lambda >0$ and the index $\beta>0$ in $\hat{K}^3$ are
chosen by a procedure called leave-one-out cross validation. 
Also $\mathscr{H}_{ K }$ is the space of functions spanned by $\{ 
 K _x: x\in \mathscr P\}$ (where $ K _x(y):= K (x,y)$) 
on a finite set $\mathscr P$ of peptides containing
$\mathscr{P}_a$. An inner product on $\mathscr{H}_{ K }$ is defined on the basis vectors as
$\left<  K _x,  K _y \right>_{\mathscr{H}_{ K }} = 
 K (x,y)$, then in general by linear extension. The norm of $f\in
\mathscr{H}_{ K }$ induced by this inner product is denoted by
$\|f\|_{ K }$. In (\ref{rls}), $f_a$ is the predicted peptide binding function.
We refer to this algorithm as ``{\bf KernelRLS}''.

For the set of HLA II alleles, with the best data available we have Table 1.
The area under the receiver operating characteristic curve
(area under the ROC curve, {\bf AUC}) is the main measure of accuracy used in the peptide binding
literature. NN-W refers to the algorithm which up to now has achieved
the most accurate results for this problem, although there are many
previous contributions as \cite{Wang2008, Lin2008, El-Manzalawy2008}. In
Section \ref{section2} there is more detail.

\begin{table}[H]
\begin{center}
\begin{tabular}
{ c | c |c c |c }
\hline\hline
\multirow{2}{*}{List of alleles, $a$ }& 
\multirow{2}{*}{$\#\mathscr{P}_a$} & \multicolumn{2}{c|}{KernelRLS} & NN-W in \cite{Nielsen2009} \\
&&RMSE&AUC&AUC\\
\hline
DRB1*0101 & 5166 & 0.18660 & \textbf{0.85707} & 0.836 \\
DRB1*0301 & 1020 & 0.18497 & \textbf{0.82813} & 0.816 \\
DRB1*0401 & 1024 & 0.24055 & \textbf{0.78431} & 0.771 \\
DRB1*0404 & 663 & 0.20702 & 0.81425 & \textbf{0.818} \\
DRB1*0405 & 630 & 0.20069 & \textbf{0.79296} & 0.781 \\
DRB1*0701 & 853 & 0.21944 & 0.83440 & \textbf{0.841} \\
DRB1*0802 & 420 & 0.19666 & \textbf{0.83538} & 0.832 \\
DRB1*0901 & 530 & 0.25398 & \textbf{0.66591} & 0.616 \\
DRB1*1101 & 950 & 0.20776 & \textbf{0.83703} & 0.823 \\
DRB1*1302 & 498 & 0.22569 & 0.80410 & \textbf{0.831} \\
DRB1*1501 & 934 & 0.23268 & \textbf{0.76436} & 0.758 \\
DRB3*0101 & 549 & 0.15945 & 0.80228 & \textbf{0.844} \\
DRB4*0101 & 446 & 0.20809 & 0.81057 & \textbf{0.811} \\
DRB5*0101 & 924 & 0.23038 & \textbf{0.80568} & 0.797 \\
\hline
Average & & 0.21100 & 0.80260 & 0.798 \\
\hline
Weighted Average & & 0.20451 & 0.82059 & 0.810\\
\hline\hline
\end{tabular}
\caption{The algorithm performance of RLS on each fixed allele in the benchmark \cite{Nielsen2009}.
If $a$ is the allele in column 1, then the number of peptides in $\mathscr{P}_a$
is given in column $2$. The root-mean-square deviation ({\bf RMSE}) scores are listed (see Section \ref{section2}).
The AUC scores of the RLS and the NN-W algorithm are listed for comparison, where a common
threshold $\theta=0.4256$ is used \cite{Nielsen2009} in the final thresholding step
into binding and non-binding (see Section \ref{subsctnFixingBetaPep} for the details). 
The best AUC in each row is marked in bold. In all the tables the weighted average scores are given 
by the weighting on the size $\#\mathscr{P}_a$ of the corresponding peptide sets $\mathscr{P}_a$.
} \label{K17}
\end{center}
\end{table}

We note the simplicity and universality of the algorithm that is based
on $\hat K^3$, which itself has this simplicity with the contributions from
the substitution matrix (i.e. BLOSUM62-2) and the sequential representation of the peptides.
There is an important generalization of the peptide binding
problem where the allele is allowed to vary. Our results on this
problem are detailed in Section \ref{section3}.

\subsection{Second Application: Clustering and Supertypes}\label{secondApplicationInIntroduction}

We consider the classification problem of DRB (HLA-DR $\beta$ chain) alleles into groups called supertypes as follows.
The understanding of DRB similarities is very important for the designation
of high population coverage vaccines. An HLA gene can generate a large number of
allelic variants and this polymorphism guarantees a population from being eradicated by an single pathogen.
Furthermore, there are no more than twelve HLA II alleles in each individual \cite{Janeway2001} and each HLA II allele binds only to specific peptides \cite{Sette1989, Yewdell1999}. As a result,
it’s difficult to design an effective vaccine for a large population.
It has been demonstrated that many HLA molecules have overlapping peptide binding sets
and there have been several attempts to group them into supertypes accordingly \cite{Sidney1996,
Sette1999, Sidney2008, Ou1998, Lund2004, Baas1999, Castelli2002}.
The supertypes are designed so that the HLA molecules in the same supertype will have a similar peptide binding specificity.

The Nomenclature Committee of the World Health Organization ({\bf WHO}) \cite{Marsh2010}
has given extensive tables on serological type assignments to DRB alleles which are
based on the works of many organizations and labs throughout the world.
In particular the HLA dictionary 2008 by Holdsworth et al. \cite{Holdsworth2009} acknowledges especially the data
from the WHO Nomenclature Committee for Factors
of the HLA system, the International Cell Exchange and the National
Marrow Donor Program. The text in Holdsworth et al., 2008 \cite{Holdsworth2009} indicates
also the ambiguities of such assignments especially in certain
serological types.

We define a set $\mathscr{N}$ of DRB alleles as follows. We downloaded 820 DRB allele sequences from the IMGT/HLA Sequence
Database \cite{Robinson2003} \footnotemark. And then 14 non-expressed alleles
were excluded and there remained 806 alleles. We use two markers ``RFL'' and ``TVQ'', each of which consists of three amino acids to identify the polymorphic part of a DRB allele. For each allele, we only consider the amino acids located between the markers ``RFL'' (the location of the first occurrence of ``RFL'') and ``TVQ'' (the location of the last occurrence of ``TVQ''). One reason is the majority
of polymorphic positions occur in exon 2 of the HLA class II genes \cite{Grody2010},
and the amino acids located between the markers ``RFL'' and ``TVQ'' constitute the whole
exon 2 \cite{Thomson2010}. The DRB alleles are encoded by 6 exons. Exon 2 is the most important component constituting an HLA II-peptide binding site. The other reason is in the HLA pseudo-sequences used in the NetMHCIIpan\cite{Nielsen2008},
all positions of the allele contacting with the peptide occur in this range.

Thus each allele is
transformed into a normal form. We should note that two different alleles
may have the same normal form. For those alleles with the same normal
form, we only consider the first one. The order is according
to the official names given by WHO. We collect the remaining 786 alleles with no duplicate
normal forms into a set, we call $\mathscr{N}$. This set not only includes all alleles listed in the tables of 
\cite{Holdsworth2009}, but also contains all new alleles since 2008 until August
2011. \footnotetext{\tt ftp://ftp.ebi.ac.uk/pub/databases/imgt/mhc/hla/DRB\_prot.fasta}

Thus $\mathscr{N}$ may be identified with a set of amino acid sequences. Next impose the kernel
$\hat{K}^3$ above on
$\mathscr{N}$ where $\beta = 0.06$, we call the kernel $\hat{K}^3_{\mathscr{N}}$.

On $\mathscr{N}$ we define a distance derived from $\hat{K}^3_{\mathscr{N}}$ by
\begin{eqnarray}
D_{L^2}(x,y) = \left( \frac{ 1 }{ \#\mathscr{N} } \sum_{z\in \mathscr{N}}
\left( \hat{K}^3_{\mathscr{N}}(x,z) -
\hat{K}^3_{\mathscr{N}}(y,z) \right)^2 \right)^{1/2} , \qquad \forall x,y\in \mathscr{N}.
\end{eqnarray}
Here and in the sequel we denote $\#A$ the size of a finite set $A$.

The DRB1*11 and DRB1*13 families of alleles have been the most
difficult to deal with by WHO and for us as well.
Therefore we will exclude the DRB1*11 and DRB1*13
families of alleles in the following cluster tree construction
with the evidence that clustering of these $2$ groups is ineffective.
They are left to be analyzed separately.\footnote{We have
found from a number of different experiments that
``they do not cluster''. Perhaps the geometric
phenomenon here is in the higher dimensional scaled topology,
i.e. the betti numbers $b_{i}>0$, for $i>0$.}

The set $\mathscr{M}$ consists of all DRB alleles except for the DRB1*11 and DRB1*13 families
of alleles. $\mathscr{M}$ is a subset of the set $\mathscr{N}$. We produce a clustering of $\mathscr{M}$ based on the $L^2$ distance $D_{L^2}$
restricted to $\mathscr{M}$, and use the OWA (Ordered Weighted Averaging)
\cite{Yager1988} based linkage instead of the ``single'' linkage in the hierarchical clustering algorithm.

This clustering uses no previous serological type information and no alignments.
We have assigned supertypes labeled ST1, ST2, ST3, ST4, ST5, ST6, ST7, ST8, ST9, ST10, ST51, ST52 and ST53 to certain clusters in the Tree shown in Figure 1 based on contents of the clusters described in Table 6. Peptides have played no role in our model. Differing from
the artificial neural network method \cite{Maiers2003, Holdsworth2009}, no ``training data'' of
any previously classified alleles are used in our clustering. We make use of the DRB amino acid sequences to build the cluster tree. Only making use of these amino acid sequences, our supertypes are in exact agreement with 
WHO assigned serological types \cite{Holdsworth2009}, as can be seen by checking the
supertypes against the clusters in Table 6.

\begin{figure}[H]
\centering
\includegraphics[width=0.9\textwidth]{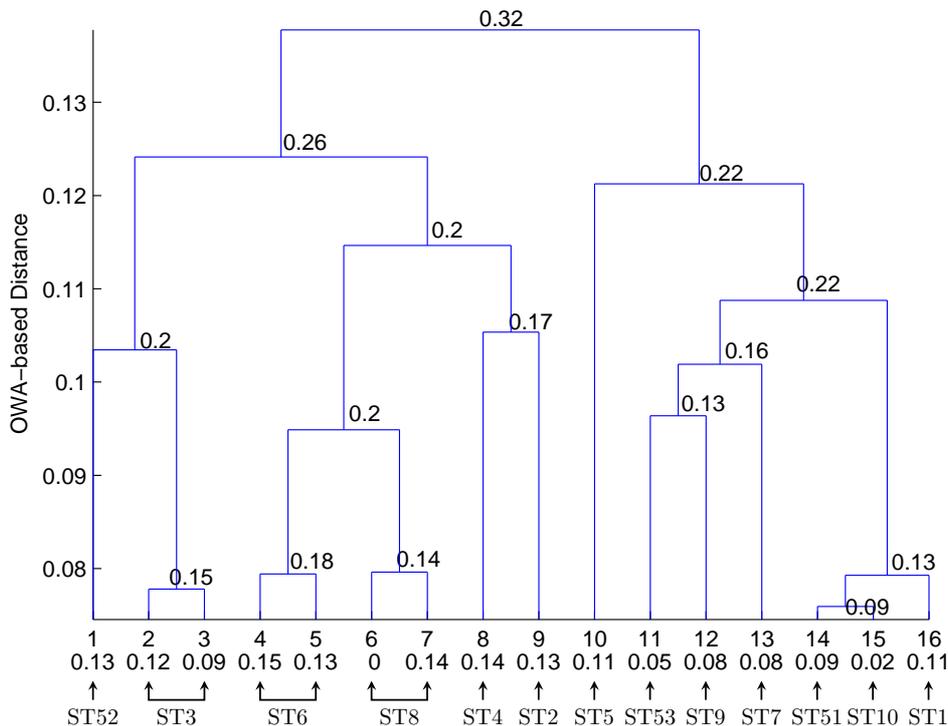}
\caption{Cluster tree on 559 DRB alleles. The diameters of the leaf nodes are given at the
bottom of the figure. The numbers given in the figure are the
diameters of the corresponding unions of clusters.}
\label{figure1}
\end{figure}

This second application is given in some detail in Section \ref{section4}.

\section{Kernel Method for Binding Affinity Prediction}\label{section2}

In this section we describe in detail the construction of our string kernel. 
The motivation is to relate the sequence information of strings (peptides or 
alleles) to their biological functions (binding affinities). A kernel works 
as a measure of similarity and supports the application of powerful machine 
learning algorithms such as RLS which we use in this paper.
For a fixed allele, binding affinity is a function on peptides with values in $[0,1]$.
The function values on some peptides are available as the data, according to which RLS 
outputs a function that predicts for a new peptide 
the binding affinity to the allele. The method is generalized in the next section to the pan-allele kernel 
algorithm that takes also the allele structure into account.

\subsection{Kernels}

We suppose throughout the paper that $X$ is a finite set. We now give the definition of a kernel,
of which an important example is our string kernel.

\begin{definition}
A symmetric function $K: X \times X \to \mathbb{R}$ is called a kernel on $X$ if it 
is positive definite, in the sense that by choosing an order 
on $X$, $K$ can be
represented as a positive definite matrix $(K(x,y))_{x,y\in X}$.
\end{definition}

Kernels have the following properties \cite{Cucker2007, Shawe-Taylor2004, Aronszajn1950}.
\begin{lemma} \label{basic}
(i) If $K$ is a kernel on $X$ then it is also a kernel on any subset $X_1$ of $X$.

(ii) If $K_1$ and $K_2$ are kernels on $X$, then $K: X 
\times X\to\mathbb{R}$ defined by $$K(x, x') = K_1(x, x') + K_2(x, x')$$ is also a kernel.

(iii) If $K_1$ is a kernel on $X_1$ and $K_2$ is a 
kernel on $X_2$, then $K: (X_1 \times X_2) \times 
(X_1 \times X_2)\to \mathbb{R}$ defined by $$K((x_1, x_2), (x'_1, x'_2)) =
K_1(x_1, x'_1) \cdot K_2(x_2, x'_2)$$ is a kernel on $X_1\times X_2$.

(iv) If $K$ is a kernel on $X$, and $f$ is a 
real-valued function on $X$ that maps no point to zero, then $K': X \times X$ defined by 
$$K'(x, x') = f(x) K(x, x') f(x')$$ is also a kernel.

(v) If $K(x, x) >0$ for all $x \in X$, then the correlation normalization $\hat{K}$ of $K$ given by
\begin{equation} \label{cor}
\hat K (x, x') = \frac{K(x,x')}{\sqrt{K(x,x)K(x',x')}}
\end{equation}
is also a kernel. 
\end{lemma}
\proof (i), (ii) and (iv) follows the definition directly. (iii) follows 
the fact that the Kronecker product of two positive definite matrices is
positive definite; see \cite{Horn1994} for details. The positive definiteness 
of a kernel $K$ guarantees that $K(x,x)>0$ for any $x$ in $X$, so (v) follows (iv).
\qed

\begin{remark}
Notice that with correlation normalization we have $\hat 
K(x,x) = 1$ for all $x \in X$. This is a desired property 
because the kernel function is usually used as a similarity 
measure, and with $\hat K$ we can say that each $x \in X$ is similar to itself.
\end{remark}

Define the real-valued function on $X$, $K_x$, by $K_x(y) = K(x,y)$.
The function space $\mathscr{H}_K = \mbox{span}\{K_x:x \in X\}$ is a Euclidean space 
with inner product $\left< K_x, K_y \right> = K(x,y)$, extended linearly to $\mathscr{H}_K$.
The norm of a function $f$ in $\mathscr{H}_K$ is denoted as $\|f\|_K$. 

\begin{remark}
The kernel can be defined even without assuming $X$ is finite; in this general case 
the kernel is referred to as a reproducing kernel \cite{Aronszajn1950}. If $X$ is finite then a
reproducing kernel is equivalent to our ``kernel''.
The theory of reproducing kernel Hilbert spaces plays an important role in learning
\cite{Cucker2007, Scholkopf2001}.
\end{remark}

On a finite set $X$ there are two notions of distance
derived from a kernel $K$. The first one is the usual distance in $\mathscr{H}_K$, that is
$$
D_{K}(x, x') = \|K_x-K_{x'}\|_K,
$$
for two points $x, x' \in X$.
The second one is the $L^2$ distance defined by
$$
D_{L^2}(x, x') = \left( \frac{1}{\# X} \sum_{t 
\in X} \left(K(x, t) - K(x', t) \right)^2 \right)^{\frac{1}{2}}.
$$

Important examples of the kernels discussed above are our kernel 
$K^3$ and its normalization $\hat{K}^3$, both defined on any finite 
$X\subset \cup_{k\geq 1} \mathscr{A}^k$

\subsection{Kernel on Strings}\label{KernelOnStrings}

We start with a finite set $\mathscr{A}$ called the alphabet. In the 
work here $\mathscr{A}$ is the set of 20 amino 
acids, but the theory
in this section applies to any other finite set. For 
example, as the name suggests, it can work on text for 
semantic analysis with a similar setting. See also \cite{Smale2010} 
for the framework in vision.

To measure a similarity among the $20$ amino acids, Henikoff and Henikoff
\cite{Henikoff1992} collect families
of related proteins, align them and find conserved regions (i.e. regions
that do not mutate frequently or greatly) as blocks in the families. The 
occurrence of each pair of amino acids in each column of every block is counted. 
A large number of occurrences indicate that in the conserved regions 
the corresponding pair of amino acids substitute each other frequently, or in another 
way of saying, that they are similar. A symmetric matrix $Q$ indexed by $\mathscr{A} \times
\mathscr{A}$ is eventually obtained by normalizing the occurrences, so that 
$\sum_{x,y\in \mathscr{A}}Q(x,y)=1$ and $Q(x,y)$
indicates the frequency of occurrences.
See \cite{Henikoff1992} for details.
The BLOSUM62 matrix is constructed accordingly.

Define $K^1: \mathscr A \times \mathscr A \to \mathbb{R}$ as
\begin{eqnarray*}
K^1(x,y) = \left( \frac{ Q(x,y) }{ p(x)p(y) } \right)^\beta, \quad \mbox{for some }\beta >0,
\end{eqnarray*}
where $p:\mathscr{A} \to [0,1]$ given by
\begin{eqnarray*}
p(x) = \sum_{y\in \mathscr A} Q(x,y),
\end{eqnarray*}
is the marginal probability distribution on $\mathscr A$.
When $\beta=1$, we name the matrix $(K^1(x,y))_{x,y\in \mathscr A}$ as
BLOSUM62-2 (one takes logarithm with base $2$, scales it with factor 2, and rounds the obtained 
matrix to integers to obtain the BLOSUM62 matrix). 
Notice that if one chooses simply $Q=\frac1m I_{m\times m}$, then one obtains the matrix 
$I_{m\times m}$ as the analogue of the BLOSUM62-2, and the corresponding $K^3$ of
the introduction is called the spectrum kernel \cite{Leslie2002}.

In matrix language $K^1$ is the Hadamard power of the BLOSUM62-2 matrix, where 
for a matrix $M = (M_{i,j})$ with positive entries and 
a number $\beta>0$, we denote $M^{\circ \beta}$ 
as the $\beta$'th Hadamard power of $M$ 
and $\log^\circ M$ as the Hadamard logarithm of $M$, and their
$(i,j)$ entries are respectively,
\begin{eqnarray*}
(M^{\circ \beta})_{i,j} := (M_{i,j})^\beta, \qquad (\log^\circ M)_{i,j}
:= \log(M_{i,j}).
\end{eqnarray*}

\begin{theorem}[Horn and Johnson\cite{Horn1994}]
Let $A$ be an $m\times m$ positive-valued symmetric matrix. The Hadamard
power $A^{\circ\beta}$ is positive definite for any $\beta>0$ if and only if
the Hadamard logarithm $\log^\circ A$ is conditionally positive definite (i.e.
positive definite on the space $V = \{ v = (v_1,\cdots, v_m)\in\mathbb{R}^m : \sum^m_{i=1}v_i = 0 \}$).
\end{theorem}

\begin{proposition}
Every positive Hadamard power of BLOSUM62-2 is positive definite. Thus the above defined $K^1$
is a kernel for every $\beta > 0$.
\end{proposition}
\proof One just shows the eigenvalues of the Hadamard logarithm on $V$
are all positive. One checks this by computer.

\begin{theorem}
Based on any kernel $K^1$, the functions $K^2_k$, $K^3$, and $\hat{K}^3$ defined 
as in the introduction are all kernels.
\end{theorem}
\proof 
The fact that $K^2_k$ is a kernel for $k\geq 1$ follows from Lemma \ref{basic} (iii).
We now prove that $K^3$ is positive definite on any finite set $X$ of strings, which then
implies the same for $\hat{K}^3$ by Lemma \ref{basic} (v). 
From Lemma \ref{basic} (i) it suffices to verify the cases that $X=X_k=\cup^k_{i=1}
\mathscr{A}^i$ for $k\geq 1$. When $k=1$, $K^3$
is just $K^1$ and hence positive definite. We assume now that $K^3$ is positive
definite on $X_k$ with $k=n$.

We claim that the matrices indexed by $X_{n+1}$,
\begin{eqnarray*}
K^3_{i, X_{n+1}} (f,g) =
\left\{
\begin{array}{cl}
\sum_{\substack{u \subset f, v\subset g \\ |u|=|v|=i }} K^2(u,v) & \mbox{if }|f|, |g|\geq i,\\
0 & \mbox{if } |f|<i \mbox{ or }|g| <i,\\
\end{array}
\right.
\end{eqnarray*}
are all positive semi-definite. In fact, for any $1\leq i\leq n$,
\begin{eqnarray}\label{K3expansion}
K^3_{i, X_{n+1}} = P_i K^2_i P_i^T,
\end{eqnarray}
where $K^2_i$ is the matrix $(K^2_i(u,v))_{u,v\in \mathscr{A}^i}$, and $P_i$ is a matrix with 
$X_{n+1}$ as the row index set and $\mathscr{A}^i$ as the column index set, and for any 
$f\in X_{n+1}$ and $u\in \mathscr{A}^i$, $P_i(f,u)$ counts the number of times $u$ occurs in $f$.
Let us explain equation (\ref{K3expansion}) a little more. For $f$ and $g$ in $X_{n+1}$,
from the definition of $P_i$ we have
\begin{eqnarray}\label{PositiveDefinitenessProof}
(P_i K^2_i P_i^T)(f,g) = \sum_{u,v\in \mathscr{A}^i}
P_i(f,u)P_i(g,v) K^2_i(u,v) = 
\sum_{\substack{u \subset f, v \subset g \\ |u|=|v|=i }}K^2_i(u,v), \quad \forall i.
\end{eqnarray}
Summing the equation (\ref{PositiveDefinitenessProof}) above over $i\in \mathbb{N}$ gives the definition of
$K^3(f,g)$.

For $i=n+1$, we have
\begin{eqnarray*}
K^3_{n+1, X_{n+1}} (f,g) = \left\{
\begin{array}{cl}
0 & f\not\in \mathscr{A}^{n+1} \mbox{ or } g\not \in \mathscr{A}^{n+1},\\
K^2_{n+1}(f,g) & \mbox{otherwise}.
\end{array}
\right.
\end{eqnarray*}
Therefore $K^3_{n+1, X_{n+1}}$ is positive definite on $\mathscr{A}^{n+1}$, 
and is zero elsewhere. Since
\begin{eqnarray*}
K^3(f,g) = \sum_{i=1}^n K^3_{i, X_{n+1}}(f,g), \quad \forall f,g\in X_n,
\end{eqnarray*}
we know that the sum of $ K^3_{i, X_{n+1}}$ with $i=1,\cdots, n$ are positive 
definite on $X_n$, and positive semi-definite on $X_{n+1}$. Because 
\begin{eqnarray*}
K^3(f,g) = \sum^{n+1}_{i=1} K^3_{i, X_{n+1}} (f,g), \quad \forall f,g\in X_{n+1},
\end{eqnarray*}
we see that $K^3$ is positive definite on $X_{n+1}$.\qed

\begin{corollary}
Our kernels $K^2_k$, $K^3$ and $\hat{K}^3$ are discriminative. That is, given any two 
strings $f,g$ in the domain of $K$, as long as $f\neq g$, we have
$D_K(f,g)>0$. Here $K$ stands for any of the three kernels.
\end{corollary}

\subsection{First Application: Peptide Affinities Prediction}\label{subsctnFixingBetaPep}

We first briefly review the RLS algorithm inspired by learning theory. 
Let $K$ be a kernel on a finite set $X$. Write 
$\mathscr{H}_K$ to denote the inner product space of functions on $X$ defined by $K$. Suppose 
$\bar{z} = \{ (x_i, y_i) \}_{i=1}^m$ is a sample set (called the training set) with $x_i \in X$
and $y_i \in \mathbb{R}$ for each $i$. The RLS uses a positive parameter 
$\lambda>0$ and $\bar{z}$ to generate the output function 
$f_{\bar{z}, \lambda}: X\to \mathbb{R}$, defined as
\begin{eqnarray}
\label{RLSalgorithm}
f_{\bar{z}, \lambda} = \mathrm{arg}\,\min_{f\in \mathscr{H}_K}\left\{
\frac1{\# \bar{z}} \sum_{(x_i, y_i)\in\bar{z}} \left( f(x_i) - y_i \right)^2
+\lambda \|f\|_K^2\right\}.
\end{eqnarray}
Since $\mathscr{H}_K$ is of finite dimension, one solves (\ref{RLSalgorithm}) by 
representing $f$ linearly by functions $K_x$ with $x\in X$ and finding the coefficients.
See \cite{Cucker2007, Scholkopf2001} for details.

\begin{remark}\label{representertheorem}
The RLS algorithm (\ref{RLSalgorithm}) is independent of the choice of the underlying space
$X$ where the function space $\mathscr{H}_K$ is defined, in the sense 
that the predicted values $f_{\bar{z},\lambda}(x)$ at $x\in X$ will not be changed if
we extend $K$ onto a large set $X'\supset X$ and re-run (\ref{RLSalgorithm})
with the same $\bar{z}$ and $\lambda$. This is guaranteed by the construction of the solution.
See, e.g. \cite{Cucker2007, Scholkopf2001}.
\end{remark} 

Five-fold cross validation is employed to evaluate the performance of the algorithms. 
Suppose $\bar{z}$ is partitioned into five divisions (we assume $m\geq 5$, which is always 
the case in this paper). Five-fold cross validation is the procedure
that validates an algorithm (with fixed parameters) as follows. We choose one of the
five divisions of the data for testing, train the algorithm on the
remaining four divisions, and predict the output function on the testing division. 
We do this test for five times so that each division is used in one time as
the testing data and thus every sample $x_i$ is labeled with both the observed value
$y_i$ and the predicted value $\tilde{y}_i$. The algorithm performance is obtained by 
comparing the two values over all the sample set. Similarly one defines the $n$-fold
cross validation for any $n\leq m$. As an important special instance, 
the $m$-fold case is also referred to as leave-one-out cross validation. 
Cross validations are also used to tune parameters.

Binding affinity measures the strength that a peptide binds to an allele, and is represented by
the IC50 score. Usually an IC50 score lies between 0 and 50,000 (nano molar).
A widely used IC50 threshold determining binding and non-binding is 500 (``binding'' if the IC50 
value is less than 500).
The bioinformatics 
community usually normalizes the scores by the function $\psi_b: (0,+\infty) \to [0,1]$ with a base $b>1$,
\begin{eqnarray}\label{IC50Normalize}
\psi_b(x) := \left\{
\begin{array}{ll}
0 & x>b,\\
1-\log_b x & 1\leq x \leq b,\\
1 & x < 1.
\end{array}
\right.
\end{eqnarray}
Without introducing any ambiguity we will in the sequel refer to the normalized 
IC50 value as the binding affinity using an appropriate value of $b$.

We test the kernel with RLS on the IEDB benchmark data set published on \cite{Nielsen2008}. 
The data set covers 14 DRB alleles, each allele $a$ with a set $\mathscr{P}_a$ of 
peptides. For any $p\in \mathscr{P}_a$, its sequence representation and the $[0,1]$-valued
binding affinity $y_{a,p}$ to the allele $a$ are both given. On this data set we compare our 
algorithm with the state-of-the-art NN-align algorithm proposed in \cite{Nielsen2009}. In \cite{Nielsen2009}
for each allele $a$, the peptide set $\mathscr{P}_a$ was divided into 5 parts for validating 
the performance\footnote{Both the data set and the 5-fold partition are available at 
{\tt http://www.cbs.dtu.dk/ suppl/immunology/NetMHCII-2.0.php}.}.

Now fix an allele $a$. Set $X = \mathscr{P} \supset \mathscr{P}_a$ (Remark \ref{representertheorem}
shows that one may select any finite $\mathscr{P}$ that contains $\mathscr{P}_a$ here). Define the kernel 
$\hat{K}^3$ on $X$ through the steps in the Introduction (leaving the power index $\beta$ to be fixed).
We use the same $5$-fold partition $\mathscr{P}_a = \cup^5_{t=1} \mathscr{P}_{a,t}$ as in \cite{Nielsen2008},
and use five-fold cross validation to test our algorithm (\ref{RLSalgorithm}) with $K = \hat{K}^3$. In the 
$t$'th test ($t=1,\cdots,5$) four parts of $\mathscr{P}_a$ are merged to be the training 
data, denoted as $\mathscr{P}^{(t)}_a = \mathscr{P}_a\backslash \mathscr{P}_{a,t}$, 
and $\mathscr{P}_{a,t}$ is left as the testing data.
For fixed $t$ and $a$, we further tune the parameter
$\beta$ in $\hat{K}^3$ and the regularization parameter $\lambda$ in (\ref{RLSalgorithm})
by leave-one-out cross validation with $\bar{z} = \mathscr{P}^{(t)}_a$.
Every pair of $\beta$ in the geometric sequence $\{ 0.001, \cdots, 10 \}$ of length 30
and $\lambda$ in the geometric sequence $\{ \mathrm{e}^{-17}, \cdots, \mathrm{e}^{-3} \}$ 
of length 15 is tested. With the optimal pair $(\beta^{(t)}_a, \lambda^{(t)}_a)$, we 
train the RLS (\ref{RLSalgorithm}) once more on $\mathscr{P}^{(t)}_a$ to give the predicted binding
function $f_{\mathscr{P}^{(t)}_a,\lambda^{(t)}_a, \beta^{(t)}_a}$ on $\mathscr{P}$. 
After the five times of testing on allele $a$, we denote
$\tilde{y}_{a,p} = f_{\mathscr{P}^{(t)}_a,\lambda^{(t)}_a, \beta^{(t)}_a} (p)$
for each $p\in\mathscr{P}_{a,t}$ and $t=1,\cdots,5$.

The RMSE score is therefore evaluated as
\begin{eqnarray*}
RMSE_a = \sqrt{
\frac1{\#\mathscr{P}_a} \sum_{p\in \mathscr{P}_a}
\left( \tilde{y}_{a,p} - y_{a,p} \right)^2.
}
\end{eqnarray*}
A smaller RMSE score indicates a better algorithm performance.
Since the affinity labels in this data set are transformed with $\psi_{b=50,000}$,
there is a threshold $\theta =\psi_{50,000}(500) \approx  0.4256$ in 
\cite{Nielsen2009} dividing the peptides $p\in \mathscr{P}_a$ into
``binding'' if $y_{a,p}>\theta$ and ``non-binding'' otherwise, to the allele $a$. Denote
$\mathscr{P}_{a,B}=\{ p\in \mathscr{P}_a: y_{a,p} > \theta \}$ and 
$\mathscr{P}_{a,N} = \mathscr{P}_a \backslash \mathscr{P}_{a,B}$. Then the AUC index
is defined to be
\begin{eqnarray}\label{AUCdef}
AUC_a = \frac{ \#\{ (p,p'): p\in \mathscr{P}_{a,B},\,
p'\in \mathscr{P}_{a,N},\, \tilde{y}_{a,p} > \tilde{y}_{a,p'}
\} }{ \left( \# \mathscr{P}_{a,B} \right)  \left( \# \mathscr{P}_{a,N} \right)} \in [0,1].
\end{eqnarray}

The sequence of ideas for each allele $a$ leads to Table \ref{K17}.
The computation also suggests a weighted optimal values of $\beta$
\begin{eqnarray}\label{betaPeptide}
\beta^*_{peptide} := \frac{ 1 }{ \sum_a\#\mathscr{P}_a } \sum_a\left\{ ( \#\mathscr{P}_a )
\left( \frac{ 1 }{ 5 }\sum^5_{t=1} \beta^{(t)}_a \right)\right\}= 0.11387.
\end{eqnarray}
 We will use this 
value in the next section.

\begin{remark}
We take the point of view that peptide binding is a matter of degree and hence is
better measured by a real number, rather than the binding--non-binding dichotomy.
Thus RMSE is a better measure than AUC.
The results in Table \ref{K17} also demonstrate that the regression-based learning
model works well.
\end{remark}

\begin{remark}
Our philosophy is that there is a kernel structure on the set of amino acid 
sequences related to their biological functions (e.g. the correspondent distances on peptides
relates to their affinities to each allele). The kernel should not depend
on the alignment information, which is a source of noise.
The performance of our kernel $\hat{K}^3$
is reflected in the modulus of continuity of the predicted
values, namely,
\begin{eqnarray*}
\Omega_a := \max_{p,p'\in\mathscr{P}_a}\frac{ |\tilde{y}_{a,p} - \tilde{y}_{a,p'}| }{ 
d(p,p')},
\end{eqnarray*}
where
\begin{eqnarray*}
d(p,p') = \| \hat{K}^3_{p} - \hat{K}^3_{p'}\|_{\hat{K}^3} = \sqrt{2 - 2\hat{K}^3(p,p')}
\end{eqnarray*}
is the distance in the space $\mathscr{H}_{\hat{K}^3}$ on peptides, and the kernel $\hat{K}^3$
is defined with $\beta = \beta^*_{peptide}$.
We list the values of $\Omega_a$ for the 14 alleles in Table \ref{omega14}. 
\begin{table}[H]
\begin{center}
\begin{tabular}
{ cc|cc|cc }
\hline\hline
Allele $a$ & $\Omega_a$ & Allele $a$ & $\Omega_a$ & Allele $a$ & $\Omega_a$  \\
\hline
DRB1*0101 & 1.2222&
DRB1*0301 & 1.0307&
DRB1*0401 & 0.9249\\
DRB1*0404 & 0.9726&
DRB1*0405 & 0.8394&
DRB1*0701 & 1.1317\\
DRB1*0802 & 0.9368&
DRB1*0901 & 0.8004&
DRB1*1101 & 0.9795\\
DRB1*1302 & 0.7745&
DRB1*1501 & 0.9843&
DRB3*0101 & 0.7395\\
DRB4*0101 & 0.8587&
DRB5*0101 & 1.0011\\
\hline\hline
\end{tabular}
\caption{The module of continuity of the predicted values.} \label{omega14}
\end{center}
\end{table}
The modulus of 
continuity can be extended to a bigger peptide set $\mathscr{P}'$ which contains the neighbourhood
of each peptide $p\in \mathscr{P}$ with respect to the metric $d$.
\end{remark}

\section{Kernel Algorithm for pan-Allele Binding Prediction}\label{section3}

We now define a pan-allele kernel on the product space of alleles and peptides.
The binding affinity data is thus a subset of this product space.
The main motivation is that by the pan-allele kernel we predict affinities to those 
alleles with few or no binding data available: this is often the case because
the MHC II alleles form a huge set (the phenomenon is often referred to as MHC II polymorphism),
and the job of determining experimentally peptide affinities to all the alleles is immense.
Also, in the pan-allele setting, one puts the binding data to different alleles together
to train the RLS. This makes the training data set larger than that was available
in the fixed allele setting, and thus helps to improve the algorithm performance.
This is verified in Table \ref{fdpa}.

Let $\mathscr{L}$ be a finite set of amino acid sequences representing the MHC II alleles.
Using a positive parameter $\beta_{allele}$ we define a kernel $\hat{K}^3_{\mathscr{L}}$
on $\mathscr{L}$ following the steps in the Introduction. Let $\mathscr{P}$ be 
a set of peptides. In the sequel we denote by $\beta_{peptide}$ specifically the
parameter used to define the kernel $\hat{K}^3_{\mathscr{P}}$ on $\mathscr{P}$. 
We define the pan-allele kernel on $\mathscr{L} \times \mathscr{P}$ as
\begin{eqnarray}\label{panKernelDefinition}
\hat{K}^3_{pan}((a,p),(a',p')) = \hat{K}^3_{\mathscr{L}}(a,a')\hat{K}^3_{\mathscr{P}}(p,p').
\end{eqnarray}
Let be given a set of data $\{(p_i, a_i, r_i)\}^m_{i=1}$.
Then for each $i$, $a_i\in \mathscr{L}$, $p_i\in \mathscr{P}$, and $r_i\in[0,1]$ is the binding 
affinity of $p_i$ to $a_i$. The RLS is applied as in Section \ref{section2}.
The output function $F: \mathscr{L}
\times\mathscr{P} \to \mathbb{R}$ is the predicted binding affinity. 

\begin{remark}
When we choose $\mathscr{L} = \{ a \}$ for a certain allele $a$, the setting and 
the algorithm reduce to the fixed-allele version studied in Section \ref{section2}.
\end{remark}

We test the pan-allele kernel with RLS (we call the algorithm 
``{\bf KernelRLSPan}'') on Nielsen's NetMHCIIpan-2.0 data set 
(we also denote by this name the algorithm published on \cite{Nielsen2010} with the data set), 
which contains 33,931 peptide-allele pairs. For peptides, amino acid sequences are given, and for
alleles, DRB names are given so that we can find out the sequence representations in $\mathscr N$
as defined in Section \ref{secondApplicationInIntroduction}.
Each pair is labeled with a 
$[0,1]$-valued binding affinity.
There are 8083 peptides and 24 alleles in $\mathscr N$ in total that appear in these peptide-allele pairs.
The whole data set is divided into 5 parts in \cite{Nielsen2010}\footnote{Both
the data set and the 5-part partition are available at {\tt http://www.cbs.dtu.dk/
suppl/immunology/NetMHCIIpan-2.0}.}.

We choose the following setting. Let $\mathscr{L} = \mathscr N$ and $\mathscr P$ be a peptide set large enough
to contain all the peptides in the data set. We use $\beta^*_{peptide}=0.11387$ as suggested in (\ref{betaPeptide})
to construct $\hat{K}^3_{\mathscr P}$ and leave the power index $\beta_{allele}$ for $\hat{K}^3_{\mathscr N}$ to be
fixed later. This defines $\hat{K}^3_{pan}$. We test the RLS algorithm by five-fold cross validation
according to the 5-part
division in \cite{Nielsen2010}. In each test we merge $4$ parts of the samples
as the training data and leave the other part as the 
testing data. Leave-one-out cross validation is further employed in 
each test to tune the parameters. We select
a pair $(\beta_{allele},\lambda)$ from the product of 
$\{ 0.02\times n : n=1,2,\cdots,8\}$ 
and $\{ \mathrm{e}^n: n=-17,-16,\cdots,-9 \}$.
The procedures are the same as used in Section \ref{subsctnFixingBetaPep} except we now 
do cross validation for the peptide-allele pairs.
In all the five tests, the pair $\beta_{allele} = 0.06$ and $\lambda = \mathrm{e}^{-13}$ achieves the 
best performance in the training data. We now use the threshold 
$\theta = \psi_{15,000}(500)\approx 0.3537$ to evaluate the AUC score, 
because the affinity values in the data set are obtained by the transform $\psi_{15,000}$.
The results of these computations are shown in Table \ref{pan-NETNielsen}.

\begin{table}[H] 
\centering\small
\begin{tabular}{l|l|cc|c} 
\hline\hline
\multirow{2}{*}{allele, $a$} & \multirow{2}{*}{\#$\mathscr{P}_a$} 
& \multicolumn{2}{|c|}{KernelRLS} & NetMHCIIpan-2.0\\
&&RMSE&AUC&AUC\\\hline
DRB1*0101 & 7685 & 0.20575 & 0.84308 & \textbf{0.846} \\ 
DRB1*0301 & 2505 & 0.18154 & 0.85095 & \textbf{0.864} \\ 
DRB1*0302 & 148  & 0.21957 & 0.71176& \textbf{0.757} \\ 
DRB1*0401 & 3116 & 0.19860 & 0.84294& \textbf{0.848} \\ 
DRB1*0404 & 577  & 0.21887 & 0.80931& \textbf{0.818} \\ 
DRB1*0405 & 1582 & 0.17459 & \textbf{0.86862}& 0.858 \\ 
DRB1*0701 & 1745 & 0.17769 & \textbf{0.87664}& 0.864 \\ 
DRB1*0802 & 1520 & 0.18732 & \textbf{0.78937}& 0.780 \\ 
DRB1*0806 & 118  & 0.23091 & 0.89214& \textbf{0.924} \\ 
DRB1*0813 & 1370 & 0.18132 & \textbf{0.88803}& 0.885 \\ 
DRB1*0819 & 116  & 0.18823 & \textbf{0.82706}& 0.808 \\ 
DRB1*0901 & 1520 & 0.19741 & \textbf{0.82220}& 0.818 \\ 
DRB1*1101 & 1794 & 0.16022 & \textbf{0.88610}& 0.883 \\ 
DRB1*1201 & 117  & 0.22740 & 0.87380& \textbf{0.892} \\ 
DRB1*1202 & 117  & 0.23322 & 0.89440 & \textbf{0.900} \\ 
DRB1*1302 & 1580 & 0.19953 & 0.82298& \textbf{0.825} \\ 
DRB1*1402 & 118  & 0.20715 & \textbf{0.86474}& 0.860 \\ 
DRB1*1404 & 30   & 0.18705 & 0.64732& \textbf{0.737} \\ 
DRB1*1412 & 116  & 0.26671 & \textbf{0.89967}& 0.894 \\ 
DRB1*1501 & 1769 & 0.19609 & \textbf{0.82858}& 0.819 \\ 
DRB3*0101 & 1501 & 0.15271 & 0.82921& \textbf{0.85} \\ 
DRB3*0301 & 160  & 0.26467 & \textbf{0.86857}& 0.853 \\ 
DRB4*0101 & 1521 & 0.16355 & \textbf{0.87138}& 0.837 \\ 
DRB5*0101 & 3106 & 0.18833 & 0.87720& \textbf{0.882} \\ \hline
Average & & 0.20035 & 0.84109 & 0.846 \\ \hline 
Weighted Average & & 0.19015 & 0.84887 & 0.849 \\ \hline\hline
\end{tabular} 
\caption{The performance of KernelRLSPan. For comparison we list the AUC scores of 
NetMHCIIpan-2.0 \cite{Nielsen2010}.
The best AUC in each row is marked in bold.}
\label{pan-NETNielsen} 
\end{table}

We implement KernelRLSPan on the fixed allele data set used in Table \ref{K17}. Recall that the data set is 
normalized with $\psi_{50,000}$ and has the five-fold division defined by \cite{Nielsen2008}.
The performance is listed in Table \ref{fdpa}, which is better 
than that of KernelRLS as listed in Table \ref{K17}.

\begin{table}[H] 
\centering\small
\begin{tabular}{lll|lll} 
\hline\hline
allele, $a$ & RMSE & AUC & allele, $a$ & RMSE & AUC\\ \hline
DRB1*0101 &  0.17650 & 0.86961&
DRB1*0301 &  0.16984 & 0.85601\\
DRB1*0401 &  0.20970 & 0.82359&
DRB1*0404 &  0.17240 & 0.88193\\
DRB1*0405 &  0.18425 & 0.84078&
DRB1*0701 &  0.17998 & 0.90231\\
DRB1*0802 &  0.16734 & 0.88496&
DRB1*0901 &  0.23562 & 0.71057\\
DRB1*1101 &  0.17073 & 0.91022&
DRB1*1302 &  0.23261 & 0.75960\\
DRB1*1501 &  0.21266 & 0.80724&
DRB3*0101 &  0.16011 & 0.79778\\
DRB4*0101 &  0.18751 & 0.84754&
DRB5*0101 &  0.18904 & 0.89585\\ \hline
\multicolumn{6}{l}{Average: \phantom{Weighted} RMSE 0.18916, AUC 0.84200}\\
\multicolumn{6}{l}{Weighted Average: RMSE 0.18496, AUC 0.85452}\\
\hline\hline
\end{tabular} 
\caption{The performance of KernelRLSPan on the fixed allele data.
For defining AUC, the transform $\psi_{50,000}$ is used as in Table \ref{K17}.}
\label{fdpa} 
\end{table}

Next, we use the whole NetMHCIIpan-2.0 data set for training, and test the algorithm performance on a 
new data set. 
A set of 64798 triples of MHC II-peptide binding data is downloaded from IEDB\footnote{
The data set was downloaded from 
{\tt http://www.immuneepitope.org/list\_page.php?
list\_type=mhc\&measured\_response=\&total\_rows=64797\&queryType=true},
on May 23, 2012.}. We pick from the set the DRB alleles,
having IC50 scores, and having explicit allele names and peptide sequences.
Those items that also appear in the NetMHCIIpan-2.0 data set are deleted. 
For the duplicated items (same peptide-allele pair and same affinity)
only one of them are kept.
All the pieces with the same peptide-allele pair yet different affinities are deleted.
We deleted those with peptide length less than $9$. (The KernelRLSPan can 
handle these peptides, while the NetMHCIIpan-2.0 cannot. The short 
peptides therefore are deleted to make
the two algorithms comparable.) For some alleles the data in the set is insufficient to define
the AUC score (i.e. the denominator in (\ref{AUCdef}) becomes zero), so we delete 
tuples containing them. Eventually we obtained 11334 peptide-allele pairs
labelled with IC50 binding affinities, which are further normalized by $\psi_{15,000}$
as in the NetMHCIIpan-2.0 data set.

Now define $\hat{K}^3_{pan}$ on 
$\mathscr{N} \times\mathscr{P}$ as in (\ref{panKernelDefinition}) with $\beta_{allele} = 0.06$ as suggested by the above computation and 
$\beta_{peptide} = 0.11387$ as suggested in (\ref{betaPeptide}).
We train on the NetMHCIIpan-2.0 data set both KernelRLSPan and NetMHCIIpan-2.0\footnote{
The code is published on {\tt http://www.cbs.dtu.dk/cgi-bin/nph-sw\_request?netMHCIIpan}.}.
In the KernelRLSPan, leave-one-out cross validation is used 
to select $\lambda$ from $\{ \mathrm{e}^{-18}, \cdots, \mathrm{e}^{-8} \}$
(the result shows that $\lambda=\mathrm{e}^{-13}$ performs the best).
The algorithm performance of the two algorithms are compared on Table \ref{newdata}.

\begin{table}[H]
\centering
\begin{tabular}{c|c|cc|cc}
\hline\hline
\multirow{2}{*}{allele, $a$} &\multirow{2}{*}{ \#$\mathscr{P}_a$ }&\multicolumn{2}{c|}{kernelRLSpan} & \multicolumn{2}{c}{NetMHCIIpan-2.0} \\
&&RMSE&AUC&RMSE&AUC\\\hline
DRB1*0101 & 1024 & 0.25519 & 0.79717 & \textbf{0.24726} & \textbf{0.82988 }\\
DRB1*0102 & 7    & \textbf{0.39748} & \textbf{0.58333} & 0.62935 & \textbf{0.58333} \\
DRB1*0103 & 41   & 0.33159 & \textbf{0.83333} & \textbf{0.32204} & \textbf{0.83333} \\
DRB1*0301 & 883  & \textbf{0.21760} & 0.80276 & 0.23975 & \textbf{0.82384} \\
DRB1*0401 & 1122 & 0.19610 & 0.79930 & \textbf{0.19363} & \textbf{0.82456} \\
DRB1*0402 & 48   & \textbf{0.23912} & \textbf{0.67321} & 0.27352 & 0.65714 \\
DRB1*0403 & 43   & 0.16381 & \textbf{0.70443} & \textbf{0.15868} & 0.66995 \\
DRB1*0404 & 494  & 0.21689 & 0.79344 & \textbf{0.20219} & \textbf{0.82517} \\
DRB1*0405 & 462  & 0.19617 & 0.78941 & \textbf{0.19387} & \textbf{0.80611} \\
DRB1*0406 & 14   & 0.19516 & 0.53846 & \textbf{0.19497} & \textbf{0.61538} \\
DRB1*0701 & 724  & 0.20853 & 0.80876 & \textbf{0.20039} & \textbf{0.84786} \\
DRB1*0801 & 24   & 0.37281 & \textbf{0.72500} & \textbf{0.34767} & 0.71250 \\
DRB1*0802 & 404  & 0.17403 & 0.80407 & \textbf{0.17181} & \textbf{0.81085} \\
DRB1*0901 & 335  & 0.21204 & 0.79524 & \textbf{0.21029} & \textbf{0.80489} \\
DRB1*1001 & 20   & 0.28082 & 0.74000 & \textbf{0.24335} & \textbf{0.92000} \\
DRB1*1101 & 811  & 0.24195 & 0.83219 & \textbf{0.23838} & \textbf{0.85071} \\
DRB1*1104 & 10   & \textbf{0.43717} & \textbf{0.76190} & 0.57082 & 0.57143 \\
DRB1*1201 & 795  & 0.25786 & \textbf{0.83178} & \textbf{0.24984} & 0.82685 \\
DRB1*1301 & 147  & \textbf{0.27014} & 0.65077 & 0.30202 & \textbf{0.70722} \\
DRB1*1302 & 499  & 0.22194 & 0.82118 & \textbf{0.21284} & \textbf{0.84258} \\
DRB1*1501 & 856  & 0.21580 & 0.83563 & \textbf{0.20869} & \textbf{0.84902} \\
DRB1*1502 & 3    & \textbf{0.13186} & \textbf{1.00000} & 0.20061 & \textbf{1.00000} \\
DRB1*1601 & 16   & 0.19556 & \textbf{0.84615} & \textbf{0.18740} & 0.76923 \\
DRB1*1602 & 12   & 0.32238 & \textbf{0.68571} & \textbf{0.30431} & 0.60000 \\
DRB3*0101 & 437  & \textbf{0.16568} & 0.74058 & 0.17860 & \textbf{0.77182} \\
DRB3*0202 & 750  & \textbf{0.16021} & 0.82543 & 0.16453 & \textbf{0.84191} \\
DRB4*0101 & 563  & \textbf{0.20594} & \textbf{0.80575} & 0.21383 & 0.78734 \\
DRB5*0101 & 774  & 0.25934 & 0.78701 & \textbf{0.25849} & \textbf{0.81950} \\
DRB5*0202 & 16   & \textbf{0.23013} & \textbf{0.71429} & 0.40554 & 0.57143 \\\hline
Average   &  &\textbf{0.24046 }& 0.76987 & 0.25947 & \textbf{0.77151} \\ \hline
Weighted Average &  & 0.21853 &0.80309 &\textbf{0.21816} & \textbf{0.82216} \\ 
\hline\hline
\end{tabular}
\caption{The performance of KernelRLSPan and NetMHCIIpan-2.0 trained on
the NetMHCIIpan-2.0 benchmark data set, tested on a new dataset 
downloaded from the IEDB. The best performance of both AUC and RMSE
scores of each row is marked in bold.}
\label{newdata}
\end{table}

In this section KernelRLSPan is tested. Tables \ref{pan-NETNielsen}, 
\ref{fdpa} and \ref{newdata}
suggest that compared with 
KernelRLS, KernelRLSPan performs much better.
Also, the kernel method uses only the substitution matrix and 
the sequence representations without direct 
alignment information but yields comparable performance with the 
state-of-the-art NetMHCIIpan-2.0 algorithm.

\section{Clustering and Supertypes}\label{section4}

In this section, we describe in detail the construction of our cluster tree and our classification of DRB alleles into supertypes. We compare the supertypes identified by our model with the serotypes designated by WHO and analyze the comparison results in detail.
\subsection{Identification of DRB Supertypes}
We classify DRB alleles into disjoint subsets by using DRB amino acid sequences and the BLOSUM62 substitution matrix. No peptide binding data or X-ray 3D structure data are used in our clustering. We obtain a classification in this way into subsets (a partition) which we call supertypes.

In Section \ref{section3}, we have defined the allele kernel on $\mathscr{N}$ as $\hat{K}^3_{\mathscr{N}}$;
the $L^2$ distance derived from $\hat{K}^3_{\mathscr{N}}$ is defined as
\begin{eqnarray*}
D_{L^2}(x,y) = \left( \frac{ 1 }{ \#\mathscr{N} } \sum_{z\in \mathscr{N}} \left( \hat{K}^3(x,z)
- \hat{K}^3(y,z)\right)^2 \right)^{1/2}, \qquad \forall x,y \in \mathscr{N}.
\end{eqnarray*}

The OWA-based linkage, defined as follows is used
to measure the proximity between clusters $X$ and $Y$ \footnotemark. Let $U=(d_{xy})_{x\in X,y\in Y}$, 
where $d_{xy}=D_{L^2}(x,y)$. After ordering (with repetitions) the elements of $U$
in descending order, we obtain an ordered vector $V = (d'_1, \ldots, d'_n), n=|U|$.
 A weighting vector $W = (w_1, \cdots, w_n)$ is associated with $V$, and the proximity between clusters $X$ and $Y$ is defined as
\begin{eqnarray*}
D_{OWA}(X,Y)= \sum_{i=1}^{n}w_id'_i.
\end{eqnarray*}
Here the weights $W$ are defined as follows \cite{Sadiq2007}:
\begin{eqnarray*}
w'_i &=& \frac{ \mathrm{e}^{i/\mu} }{ \mu }, \quad i=1,2,\cdots, n,\\
w_i &=& \frac{ w'_i }{ \sum_{j=1}^n w'_j }, \quad i = 1,2, \cdots, n,
\end{eqnarray*}
where $\mu = \gamma(1+n)$ and $\gamma$ is chosen appropriately as 0.1. This weighting gives
more importance to pairs $(x,y)$ which have smaller distance. \footnotetext{Another way
of measuring distance between clusters is the Hausdorff distance.}

Hierarchical clustering \cite{day1984efficient} is applied to build a cluster tree. A cluster tree is a tree on which every node represents the cluster of the set of all leaves descending from that node. The $L_2$ distance $D_{L^2}$ is used to measure the distance between alleles $x$ and $y$, $x,y\in \mathscr{M}$ and OWA-based linkage is used to measure the proximity between clusters $X$ and $Y$, $X, Y \subseteq\mathscr{M}$ instead of ``single'' linkage. This algorithm is a bottom-up approach.
At the beginning, each allele is treated as a singleton cluster,
and then successively one merges two nearest clusters $X$ and $Y$ into a union
cluster, the process stopping when all unions of clusters have
been merged into a single cluster.

This cluster tree, associated to $\mathscr{M}$, has thus 559 leaves. We cut the cluster tree at 16 clusters, an appropriate level to separate different families of alleles. The upper part of this tree is shown in Figure 1. The contents of the clusters are given in Table 6. We assign supertypes to certain clusters in the cluster tree based on the contents of the clusters described in Table 6. A supertype is based on one or two clusters in Table 6. If two clusters in Table 6 are closest in the tree, and the alleles in which are in the same family, they are assigned an identical supertype. Thirteen
supertypes are defined in this way, which we name ST1, ST2, ST3,
ST4, ST5, ST6, ST7, ST8, ST9, ST10, ST51, ST52 and ST53.
The corresponding cluster diameters are 0.11, 0.13, 0.15, 0.14, 0.11, 0.18, 0.08,
0.14, 0.08, 0.02, 0.09, 0.13 and 0.05, respectively.

The diameter of a cluster $Z$ is defined as
\begin{eqnarray}
diameter(Z)=\max_{x,y\in Z}D_{L^2}(x,y).
\end{eqnarray}

The DRB alleles in the first ten supertypes are gathered from the DRB1 locus. The DRB alleles in the ST51, ST52 and ST53 supertypes are gathered from the DRB5 , DRB3 and DRB4 loci, respectively.

\subsection{Serotype designation of HLA-DRB alleles}
There is a historically developed classification, based on extensive works of medical labs and organizations, that groups alleles into what are called serotypes. This classification is oriented to immunology and diseases associated to gene variation in humans. It uses peptide binding data, 3D structure, X-ray diffraction and other tools. When the confidence level is sufficiently high, WHO assigns a serotype to an allele as in Table 6 where a number prefixed by DR follows the name of that allele.

There are four DRB genes (DRB1/DRB3/DRB4/DRB5) in the
HLA-DRB region \cite{Janeway2001}. The DRB1 gene/locus is much more polymorphic
than the DRB3/DRB4/DRB5 genes/loci \cite{Bittar1997}. More than 800 allelic
variants are derived from the exon 2 of the DRB genes in humans
\cite{Galan2010}. The WHO Nomenclature Committee for Factors of the HLA
System assigns an official name for each identified allele
sequence, e.g. DRB1*01:01. The characters before the separator ``*'' describe
the name of the gene, the first two digits correspond to the
allele family and the third and fourth digits correspond to
a specific HLA protein. See Table 6 for examples of how the alleles are named. If two HLA alleles belong to the same
family, they often correspond to the same serological antigen,
and thus the first two digits are meant to suggest serological types. So for those alleles which are not assigned serotypes by WHO, WHO has suggested serotypes for them according to their official names or allele families.
\subsection{Comparison of identified supertypes to designated sero\-types}
In Section 4.1, we have identified thirteen supertypes and in Section 4.2 we have introduced the WHO assigned serotypes. In the following, we compare these two classifications.

By using the cluster tree given in Figure 1 and the contents of the clusters described in Table 6, we have named our supertypes with prefix ``ST'' paralleled to the serotype names. The detailed information of DRB alleles
and serological types for these 13 supertypes is given
in Table 6. Our supertype clustering recovers the WHO serotype classification and provides further insight into the classification of DRB alleles which are not assigned serotypes. There are 559 DRB alleles in Table 6, and only 138 DRB
alleles have WHO assigned serotypes. Table 7 gives the relationship between the broad serological types and the split serological types. As shown in Tables 6 and 7, our supertypes assigned
to these 138 DRB alleles are in exact agreement with the WHO
assigned broad serological types (see Table 7). Extensive medical/biological information was used by WHO to assign serological type whereas solely DRB amino acid sequences were used in our supertype clustering. All alleles with WHO assigned DR52, DR3, DR6, DR8, DR4, DR2, DR5, DR53, DR9, DR7, DR51, DR10 and DR1-serotype are classified, respectively, into the ST52, ST3, ST6, ST8, ST4, ST2, ST5, ST53, ST9, ST7, ST51, ST10 and ST1-supertype. The other 461 alleles in Table 6 are not assigned serotypes by WHO in \cite{Holdsworth2009}. However, WHO has suggested serotypes for them according to their official names or allele families; that is, if two DRB alleles are in the same family, they belong to the same serotype.  Our clustering confirms that this suggestion is reasonable, as can be checked from the clusters in Table 6.

We make some remarks on Figure 1 and Table 6 as follows.

ST52: This supertype consists of exactly the DRB3 alleles with the exception of DRB1*0338 (a new allele and unassigned by WHO \cite{Holdsworth2009}).

ST3: This supertype consists of cluster 2 and cluster 3 in the cluster tree and contains
63 DRB1*03 alleles with two exceptions: DRB3*0115 and DRB1*1525. The DRB3*0115 is grouped with the DRB1*03 alleles
in a number of different experiments done by us, and the
DRB1*1525 is a new allele and unassigned
by WHO. Here, the DR3-serotype is a broad serotype which consists of three split serotypes, DR3, DR17 and DR18 (see Table 7).

ST6: This supertype consists of cluster 4 and cluster 5 and consists
of exactly 102 DRB1*14 alleles. Here, the DR6-serotype is a broad serotype which consists of five split serotypes, DR6, DR13, DR14, DR1403 and DR1404.

ST8: This supertype consists of cluster 6 and cluster 7 and mainly contains 46 DRB1*08 alleles (The serological designation of DRB1*1415 is
DR8 by WHO.). The unassigned alleles DRB1*1425, DRB1*1440, DRB1*1442, DRB1*1469, DRB1*1477 and DRB1*1484 are DRB1*14 alleles, but they are classified into the ST8 supertype. Both DRB1*14116
and DRB1*14102 are new allele sequences that do not
exist in the tables of \cite{Holdsworth2009, Marsh2010} and they are classified into the ST8 supertype too.

Supertypes 52, 4, 2, 5, 53, 9, 7, 51, 10 and 1 correspond, respectively, to clusters 1, 8, 9, 10, 11, 12, 13, 14, 15 and 16 in the cluster tree.

ST4: This supertype consists of exactly 99 DRB1*04 alleles.

ST2: This supertype consists of 53 DRB1*15 alleles and 16 DRB1*16 alleles. Here, the DR2-serotype is a broad serotype which consists of three split serotypes, DR2, DR15 and DR16.

ST5: This supertype contains exactly 29 DRB1*12 alleles. The DRB1*0832 is undefined
by experts in \cite{Holdsworth2009}, but its serological
designation by the neural network algorithm \cite{Maiers2003} is DR8 or DR12. We classify it into the ST5 supertype. The DR5-serotype is a broad serotype which consists of two split serotypes, DR11 and DR12.

ST53: This supertype consists of exactly the DRB4 alleles.

ST9: This supertype contains exactly the DRB1*09 alleles with the exception of DRB5*0112. The DRB5*0112 is
undefined by experts in \cite{Holdsworth2009}. And from a
number of different experiments done by us, DRB5*0112 is clustered with the DRB1*09 family of alleles.

ST7: This supertype consists of exactly 19 DRB1*07 alleles.

ST51: This supertype consists of exactly 15 DRB5 alleles.

ST10: This supertype is the smallest supertype and consists of exactly 3 DRB1*10 alleles.

ST1: This supertype consists of exactly 36 DRB1*01 alleles. Here, the DR1-serotype is a broad serotype which consists of two split serotypes, DR1 and DR103.

For the DRB alleles, there are thirteen broad serotypes given by WHO, and our clustering classifies all alleles which are assigned the same broad serotype to the same supertype. And for the alleles which are not assigned serotypes, our supertypes confirm the nomenclature of WHO.

As can be seen from Figure 1, the ST52 supertype is closest to the ST3 supertype. The ST53 supertype is closest to the ST9 and ST7 supertypes. The ST51 supertype is closest to the ST10 and ST1 supertypes.

\subsection{Previous work in perspective}
In 1999, Sette and Sidney asserted that all HLA I alleles can be classified
into nine supertypes \cite{Sette1999, Sidney2008}. This classification is defined
based on the structural motifs derived from experimentally determined
binding data. The alleles in the same supertype comprise the same
peptide binding motifs and bind to largely overlapping sets of
peptides. Essentially, the supertype classification problem is
to identify peptides that can bind to a group of HLA molecules.
Besides many works on HLA class I supertype classification, some
works have been proposed to identify supertypes for HLA class II.
In 1998, through analyzing a large set of biochemical synthetic peptides and a panel of HLA-DR binding assays, Southwood et al. \cite{southwood1998several} asserted that seven common HLA-DR alleles, e.g. DRB1*0101, DRB1*0401, DRB1*0701, DRB1*0901, DRB1*1302, DRB1*1501 and DRB5*0101 had similar peptide binding specificity and should be grouped into one supertype. By the use of HLA ligands, Lund et al. \cite{Lund2004} clustered 50 DRB alleles into nine supertypes by a Gibbs Sampling algorithm. Both of these studies used peptide binding data and this resulted in the limited number of DRB alleles available for classification. The work of Doytchinova and Flower \cite{Doytchinova2005}, classified 347 DRB alleles into 5 supertypes by the use of both protein sequences and 3D structural data. Ou et al. \cite{Ou1998}. defined seven supertypes based on similarity of function rather than on sequence or structure. To our knowledge, our study is the first to identify HLA-DR supertypes solely based on DRB amino acid sequence data.

\newcommand\CaptionLabelFont{\small}
\newcommand\CaptionTextFont{\small}
\begin{center}
\footnotesize
    \begin{longtable}{l|llllll}

    \hline\hline

    \textbf{Super-} & \textbf{Allele} & \textbf{Sero-} & \textbf{Allele} & \textbf{Sero-} & \textbf{Allele} & \textbf{Sero-} \\
    \textbf{type} & &\textbf{type} & & \textbf{type}&&\textbf{type}\\
    \hline
    \textbf{ST52} & \textbf{Cluster 1} &       &       &       &       &  \\
          & DRB3*0101(2)        & DR52   & DRB3*0108(U.)       & --      & DRB3*0212(U.)       & --      \\
          & DRB3*0106(s.s.)     & DR52   & DRB3*0102(s.s.)     & --      & DRB3*0226           & --      \\
          & DRB3*0110(s.s.)     & DR52   & DRB3*0112           & --      & DRB3*0222(U.)       & --      \\
          & DRB3*0301           & DR52   & DRB3*0105(U.)       & --      & DRB3*0204(U.)       & --      \\
          & DRB3*0209           & DR52   & DRB3*0103(s.s.)(U.) & --      & DRB3*0213(U.)       & --      \\
          & DRB3*0302(s.s.)     & DR52   & DRB3*0113           & --      & DRB3*0215(U.)       & --      \\
          & DRB3*0107(s.s.)     & DR52   & DRB3*0111(U.)       & --      & DRB3*0218(U.)       & --      \\
          & DRB3*0203(s.s.)     & DR52   & DRB3*0114           & --      & DRB3*0205(U.)       & --      \\
          & DRB3*0211           & DR52   & DRB3*0303           & --      & DRB3*0225           & --      \\
          & DRB3*0201(2)        & DR52   & DRB3*0109(U.)       & --      & DRB3*0219(U.)       & --      \\
          & DRB3*0202(2)        & DR52   & DRB3*0206(s.s.)     & --      & DRB3*0216(U.)       & --      \\
          & DRB3*0210           & DR52   & DRB3*0220(U.)       & --      & DRB3*0221(U.)       & --      \\
          & DRB3*0208(s.s.)     & DR52   & DRB3*0223           & --      & DRB3*0227           & --      \\
          & DRB3*0207(s.s.)     & DR52   & DRB3*0217(U.)       & --      &       &  \\
          & DRB1*0338           & --      & DRB3*0214(U.)       & --      &       &  \\
          \hline
    \textbf{ST3} & \textbf{Cluster 2} &       &       &       &       &  \\
          & DRB1*0323       & DR3    & DRB1*0334       & --      & DRB1*0358       & --      \\
          & DRB1*0301(2)    & DR17   & DRB1*0364       & --      & DRB1*0308       & --      \\
          & DRB1*0305       & DR3    & DRB1*0361       & --      & DRB1*0326       & --      \\
          & DRB1*0311       & DR17   & DRB1*0332       & --      & DRB1*0313       & --      \\
          & DRB1*0304       & DR17   & DRB1*0328       & --      & DRB1*0360       & --      \\
          & DRB1*0306       & DR3    & DRB1*0362       & --      & DRB1*0324       & --      \\
          & DRB1*0307       & DR3    & DRB1*0346       & --      & DRB1*0352       & --      \\
          & DRB1*0314       & DR3    & DRB1*0336       & --      & DRB1*0365       & --      \\
          & DRB1*0315       & DR3    & DRB1*0357       & --      & DRB1*0329       & --      \\
          & DRB1*0312(s.s.) & DR3    & DRB1*0339       & --      & DRB1*0327       & --      \\
          & DRB1*0302       & DR18   & DRB1*0333       & --      & DRB1*0353       & --      \\
          & DRB1*0303       & DR18   & DRB1*0319       & --      & DRB1*0321       & --      \\
          & DRB1*0310       & DR17   & DRB1*0348       & --      & DRB1*0343       & --      \\
          & DRB1*0342       & --      & DRB1*0363       & --      & DRB1*0330       & --      \\
          & DRB1*0345       & --      & DRB1*0322       & --      & DRB1*0325       & --      \\
          & DRB1*0355       & --      & DRB1*0309       & --      & DRB1*0344       & --      \\
          & DRB1*0359       & --      & DRB1*0337       & --      & DRB1*0331       & --      \\
          & DRB1*0354       & --      & DRB1*0351       & --      & DRB1*0335       & --      \\
          & DRB1*0320       & --      & DRB1*0347       & --      & DRB3*0115       & --      \\
          & DRB1*0356       & --      & DRB1*0318       & --      & DRB1*0316(s.s.) & --      \\\hhline{~------}
          & \textbf{Cluster 3} &       &       &       &       &  \\
          & DRB1*1525 & --     & DRB1*0340 & --     & DRB1*0317 & -- \\
          & DRB1*0349 & --     & DRB1*0341 & --     &       &  \\
          \hline
    \textbf{ST6} & \textbf{Cluster 4} &       &       &       &       &  \\
          & DRB1*1410    & DR14   & DRB1*1482    & --      & DRB1*1472    & --      \\
          & DRB1*1401(4) & DR14   & DRB1*1462    & --      & DRB1*14101   & --      \\
          & DRB1*1426    & DR14   & DRB1*1470    & --      & DRB1*1434    & --      \\
          & DRB1*1407    & DR14   & DRB1*1438    & --      & DRB1*1423    & --      \\
          & DRB1*1460    & DR14   & DRB1*14112   & --      & DRB1*1445    & --      \\
          & DRB1*1450    & DR14   & DRB1*1490    & --      & DRB1*1443    & --      \\
          & DRB1*1404    & DR1404 & DRB1*1486    & --      & DRB1*1456    & --      \\
          & DRB1*1449    & DR14   & DRB1*1497    & --      & DRB1*14103   & --      \\
          & DRB1*1411    & DR14   & DRB1*1435    & --      & DRB1*1444    & --      \\
          & DRB1*1408    & DR14   & DRB1*1455    & --      & DRB1*1496    & --      \\
          & DRB1*1414    & DR14   & DRB1*1431    & --      & DRB1*14100   & --      \\
          & DRB1*1405    & DR14   & DRB1*1493    & --      & DRB1*1436    & --      \\
          & DRB1*1420    & DR14   & DRB1*1428    & --      & DRB1*1465    & --      \\
          & DRB1*1422    & DR14   & DRB1*1471    & --      & DRB1*1464    & --      \\
          & DRB1*1416    & DR6    & DRB1*1468    & --      & DRB1*1495    & --      \\
          & DRB1*1439    & --      & DRB1*1432    & --      & DRB1*1459    & --      \\
          & DRB1*1499    & --      & DRB1*14111   & --      & DRB1*1491    & --      \\
          & DRB1*1461    & --      & DRB1*14104   & --      & DRB1*1441    & --      \\
          & DRB1*14117   & --      & DRB1*1458    & --      & DRB1*1437    & --      \\
          & DRB1*1487    & --      & DRB1*1473    & --      & DRB1*1457    & --      \\
          & DRB1*1475    & --      & DRB1*1479    & --      & DRB1*14105   & --      \\
          & DRB1*1488    & --      & DRB1*14107   & --      & DRB1*1474    & --      \\
          & DRB1*14110   & --      & DRB1*1476    & --      &       &  \\\hhline{~------}
          & \textbf{Cluster 5} &       &       &       &       &  \\
          & DRB1*1419  & DR14   & DRB1*1452  & --      & DRB1*1433  & --      \\
          & DRB1*1402  & DR14   & DRB1*14108 & --      & DRB1*1424  & --      \\
          & DRB1*1429  & DR14   & DRB1*1483  & --      & DRB1*14109 & --      \\
          & DRB1*1406  & DR14   & DRB1*1481  & --      & DRB1*14115 & --      \\
          & DRB1*1418  & DR6    & DRB1*1494  & --      & DRB1*1467  & --      \\
          & DRB1*1413  & DR14   & DRB1*1447  & --      & DRB1*1498  & --      \\
          & DRB1*1421  & DR14   & DRB1*1451  & --      & DRB1*1463  & --      \\
          & DRB1*1417  & DR6    & DRB1*14106 & --      & DRB1*1485  & --      \\
          & DRB1*1427  & DR14   & DRB1*1489  & --      & DRB1*1478  & --      \\
          & DRB1*1403  & DR1403 & DRB1*1430  & --      & DRB1*1448  & --      \\
          & DRB1*1412  & DR14   & DRB1*1409  & --      &       &  \\
          & DRB1*1446  & --      & DRB1*1480  & --      &       &  \\
          \hline
    \textbf{ST8} & \textbf{Cluster 6} &       &       &       &       &  \\
          & DRB1*1442(U.) & --     &       &       &       &  \\\hhline{~------}
          & \textbf{Cluster 7} &       &       &       &       &  \\
          & DRB1*0809     & DR8    & DRB1*1477     & --      & DRB1*0808     & --      \\
          & DRB1*1415     & DR8    & DRB1*1440     & --      & DRB1*0844     & --      \\
          & DRB1*0814     & DR8    & DRB1*1484     & --      & DRB1*0835     & --      \\
          & DRB1*0812     & DR8    & DRB1*0846     & --      & DRB1*0836     & --      \\
          & DRB1*0803     & DR8    & DRB1*0848     & --      & DRB1*0847     & --      \\
          & DRB1*0810     & DR8    & DRB1*0819     & --      & DRB1*0825     & --      \\
          & DRB1*0817     & DR8    & DRB1*0827     & --      & DRB1*0834     & --      \\
          & DRB1*0811     & DR8    & DRB1*0829     & --      & DRB1*0828     & --      \\
          & DRB1*0801     & DR8    & DRB1*0837     & --      & DRB1*0845     & --      \\
          & DRB1*0807     & DR8    & DRB1*0839     & --      & DRB1*0830     & --      \\
          & DRB1*0806     & DR8    & DRB1*0822     & --      & DRB1*0824     & --      \\
          & DRB1*0805     & DR8    & DRB1*0815     & --      & DRB1*0820(U.) & --      \\
          & DRB1*0818     & DR8    & DRB1*0840     & --      & DRB1*14116    & --      \\
          & DRB1*0816     & DR8    & DRB1*0838     & --      & DRB1*14102    & --      \\
          & DRB1*0802     & DR8    & DRB1*0826     & --      & DRB1*0842     & --      \\
          & DRB1*0804     & DR8    & DRB1*0843     & --      & DRB1*0841     & --      \\
          & DRB1*0813     & DR8    & DRB1*0833     & --      & DRB1*1425     & --      \\
          & DRB1*0821     & --      & DRB1*0823     & --      & DRB1*1469     & --      \\
          \hline
    \textbf{ST4} & \textbf{Cluster 8} &       &       &       &       &  \\
          & DRB1*0420(s.s.) & DR4    & DRB1*0438       & --      & DRB1*0490       & --      \\
          & DRB1*0401       & DR4    & DRB1*0434       & --      & DRB1*0487       & --      \\
          & DRB1*0464       & DR4    & DRB1*0475       & --      & DRB1*0430       & --      \\
          & DRB1*0408       & DR4    & DRB1*0476       & --      & DRB1*0448       & --      \\
          & DRB1*0416       & DR4    & DRB1*0472       & --      & DRB1*0467       & --      \\
          & DRB1*0426       & DR4    & DRB1*0435       & --      & DRB1*0483       & --      \\
          & DRB1*0442       & DR4    & DRB1*0443       & --      & DRB1*0480       & --      \\
          & DRB1*0432(s.s.) & DR4    & DRB1*0479       & --      & DRB1*0462       & --      \\
          & DRB1*0423       & DR4    & DRB1*0440       & --      & DRB1*0457       & --      \\
          & DRB1*0404       & DR4    & DRB1*0470       & --      & DRB1*0497       & --      \\
          & DRB1*0413       & DR4    & DRB1*0444       & --      & DRB1*0463       & --      \\
          & DRB1*0431       & DR4    & DRB1*0456       & --      & DRB1*0498       & --      \\
          & DRB1*0403       & DR4    & DRB1*0455       & --      & DRB1*0449       & --      \\
          & DRB1*0407(2)    & DR4    & DRB1*0433       & --      & DRB1*04102      & --      \\
          & DRB1*0429       & DR4    & DRB1*0439       & --      & DRB1*0441       & --      \\
          & DRB1*0424       & DR4    & DRB1*0460       & --      & DRB1*0446       & --      \\
          & DRB1*0409       & DR4    & DRB1*0450       & --      & DRB1*0485       & --      \\
          & DRB1*0405       & DR4    & DRB1*0496       & --      & DRB1*0478       & --      \\
          & DRB1*0410       & DR4    & DRB1*0451       & --      & DRB1*0465       & --      \\
          & DRB1*0428       & DR4    & DRB1*0471       & --      & DRB1*0491       & --      \\
          & DRB1*0417       & DR4    & DRB1*04100      & --      & DRB1*0468       & --      \\
          & DRB1*0411       & DR4    & DRB1*0488       & --      & DRB1*0477       & --      \\
          & DRB1*0422       & DR4    & DRB1*0493       & --      & DRB1*0484       & --      \\
          & DRB1*0406       & DR4    & DRB1*0427       & --      & DRB1*0447       & --      \\
          & DRB1*0421       & DR4    & DRB1*0452       & --      & DRB1*0436       & --      \\
          & DRB1*0419       & DR4    & DRB1*04101      & --      & DRB1*0454       & --      \\
          & DRB1*0425(s.s.) & DR4    & DRB1*0474       & --      & DRB1*0437       & --      \\
          & DRB1*0414       & DR4    & DRB1*0495       & --      & DRB1*0453       & --      \\
          & DRB1*0402       & DR4    & DRB1*0459       & --      & DRB1*0418       & --      \\
          & DRB1*0415       & DR4    & DRB1*0473       & --      & DRB1*0458       & --      \\
          & DRB1*0499       & --      & DRB1*0461       & --      & DRB1*0486       & --      \\
          & DRB1*0482       & --      & DRB1*0445       & --      & DRB1*0412       & --      \\
          & DRB1*0466       & --      & DRB1*0489       & --      & DRB1*0469       & --      \\
          \hline
    \textbf{ST2} & \textbf{Cluster 9} &       &       &       &       &  \\
          & DRB1*1501(2)    & DR15   & DRB1*1533       & --      & DRB1*1548       & --      \\
          & DRB1*1505       & DR15   & DRB1*1553       & --      & DRB1*1512       & --      \\
          & DRB1*1506       & DR15   & DRB1*1524       & --      & DRB1*1515       & --      \\
          & DRB1*1503       & DR15   & DRB1*1509       & --      & DRB1*1557       & --      \\
          & DRB1*1508       & DR2    & DRB1*1549       & --      & DRB1*1511       & --      \\
          & DRB1*1502(2)    & DR15   & DRB1*1541       & --      & DRB1*1538       & --      \\
          & DRB1*1504       & DR15   & DRB1*1540       & --      & DRB1*1529       & --      \\
          & DRB1*1507       & DR15   & DRB1*1523       & --      & DRB1*1545       & --      \\
          & DRB1*1602       & DR16   & DRB1*1518       & --      & DRB1*1554       & --      \\
          & DRB1*1605(s.s.) & DR16   & DRB1*1537       & --      & DRB1*1510       & --      \\
          & DRB1*1601       & DR16   & DRB1*1514       & --      & DRB1*1521       & --      \\
          & DRB1*1609       & DR16   & DRB1*1544       & --      & DRB1*1612       & --      \\
          & DRB1*1603       & DR2    & DRB1*1526       & --      & DRB1*1617       & --      \\
          & DRB1*1604       & DR16   & DRB1*1539       & --      & DRB1*1611       & --      \\
          & DRB1*1528       & --      & DRB1*1530       & --      & DRB1*1614       & --      \\
          & DRB1*1535       & --      & DRB1*1531       & --      & DRB1*1618       & --      \\
          & DRB1*1532       & --      & DRB1*1556       & --      & DRB1*1610       & --      \\
          & DRB1*1542       & --      & DRB1*1555       & --      & DRB1*1608       & --      \\
          & DRB1*1551       & --      & DRB1*1516       & --      & DRB1*1615       & --      \\
          & DRB1*1552       & --      & DRB1*1522       & --      & DRB1*1607       & --      \\
          & DRB1*1536       & --      & DRB1*1546       & --      & DRB1*1616       & --      \\
          & DRB1*1520       & --      & DRB1*1547       & --      & DRB1*1527       & --      \\
          & DRB1*1543       & --      & DRB1*1558       & --      & DRB1*1534       & --      \\
          \hline
    \textbf{ST5} & \textbf{Cluster 10} &       &       &       &       &  \\
          & DRB1*1202     & DR12   & DRB1*1215     & --      & DRB1*1230     & --      \\
          & DRB1*1201(4)  & DR12   & DRB1*1219     & --      & DRB1*1207     & --      \\
          & DRB1*1203     & DR12   & DRB1*1216     & --      & DRB1*1229     & --      \\
          & DRB1*1205     & DR12   & DRB1*1221     & --      & DRB1*1234     & --      \\
          & DRB1*1220     & --      & DRB1*1208     & --      & DRB1*1222     & --      \\
          & DRB1*1233     & --      & DRB1*1212     & --      & DRB1*1223     & --      \\
          & DRB1*1218     & --      & DRB1*1225     & --      & DRB1*1227     & --      \\
          & DRB1*1213     & --      & DRB1*1211     & --      & DRB1*1209     & --      \\
          & DRB1*1232     & --      & DRB1*1228     & --      & DRB1*1204     & --      \\
          & DRB1*1226     & --      & DRB1*1214     & --      & DRB1*0832(U.) & --      \\
          \hline
    \textbf{ST53} & \textbf{Cluster 11} &       &       &       &       &  \\
          & DRB4*0101(3)        & DR53   & DRB4*0104(U.)       & --      & DRB4*0107(U.)       & --      \\
          & DRB4*0105(s.s.)     & DR53   & DRB4*0102(s.s.)(U.)          & --      & DRB4*0108 & --      \\
          \hline
    \textbf{ST9} & \textbf{Cluster 12} &       &       &       &       &  \\
          & DRB1*0901     & DR9    & DRB1*0912     & --      & DRB1*0915     & --      \\
          & DRB1*0905     & DR9    & DRB1*0906     & --      & DRB1*0911     & --      \\
          & DRB1*0910     & --      & DRB1*0908     & --      & DRB1*0914     & --      \\
          & DRB1*0916     & --      & DRB1*0904     & --      & DRB5*0112(U.) & --      \\
          & DRB1*0907     & --      & DRB1*0903     & --      & DRB1*0902     & --      \\
          & DRB1*0909     & --      & DRB1*0913     & --      &       &  \\
          \hline
    \textbf{ST7} & \textbf{Cluster 13} &       &       &       &       &  \\
          & DRB1*0703 & DR7    & DRB1*0721 & --      & DRB1*0708 & --      \\
          & DRB1*0701 & DR7    & DRB1*0716 & --      & DRB1*0711 & --      \\
          & DRB1*0709 & DR7    & DRB1*0713 & --      & DRB1*0717 & --      \\
          & DRB1*0704 & DR7    & DRB1*0714 & --      & DRB1*0707 & --      \\
          & DRB1*0715 & --      & DRB1*0712 & --      & DRB1*0706 & --      \\
          & DRB1*0719 & --      & DRB1*0720 & --      &       &  \\
          & DRB1*0705 & --      & DRB1*0718 & --      &       &  \\
          \hline
    \textbf{ST51} & \textbf{Cluster 14} &       &       &       &       &  \\
          & DRB5*0101           & DR51   & DRB5*0104(U.)       & --      & DRB5*0106(U.)       & --      \\
          & DRB5*0102           & DR51   & DRB5*0103(U.)       & --      & DRB5*0111(U.)       & --      \\
          & DRB5*0107(s.s.)     & DR51   & DRB5*0113(U.)       & --      & DRB5*0204(U.)       & --      \\
          & DRB5*0202           & DR51   & DRB5*0109(s.s.)(U.) & --      & DRB5*0203(U.)       & --      \\
          & DRB5*0105(U.)       & --      & DRB5*0114           & --      & DRB5*0205(U.)       & --      \\
          \hline
    \textbf{ST10} & \textbf{Cluster 15} &       &       &       &       &  \\
          & DRB1*1001 & DR10   & DRB1*1003 & --      & DRB1*1002 & --      \\
          \hline
    \textbf{ST1} & \textbf{Cluster 16} &       &       &       &       &  \\
          & DRB1*0107 & DR1    & DRB1*0120 & --      & DRB1*0135 & --      \\
          & DRB1*0101 & DR1    & DRB1*0127 & --      & DRB1*0111 & --      \\
          & DRB1*0102 & DR1    & DRB1*0112 & --      & DRB1*0117 & --      \\
          & DRB1*0104 & DR1    & DRB1*0128 & --      & DRB1*0118 & --      \\
          & DRB1*0109 & DR1    & DRB1*0136 & --      & DRB1*0115 & --      \\
          & DRB1*0103 & DR103  & DRB1*0131 & --      & DRB1*0106 & --      \\
          & DRB1*0113 & DR1    & DRB1*0132 & --      & DRB1*0126 & --      \\
          & DRB1*0122 & --      & DRB1*0119 & --      & DRB1*0137 & --      \\
          & DRB1*0124 & --      & DRB1*0130 & --      & DRB1*0123 & --      \\
          & DRB1*0110 & --      & DRB1*0121 & --      & DRB1*0108 & --      \\
          & DRB1*0129 & --      & DRB1*0105 & --      & DRB1*0114 & --      \\
          & DRB1*0134 & --      & DRB1*0125 & --      & DRB1*0116 & --      \\
    \hline\hline
    \multicolumn{7}{c}{\phantom{\textbf{DRB3/4/5 serological families}}}  \\
    \caption{Overview of clusters of HLA-DR alleles with split serological types assigned by
    WHO.}\label{Bigtable}\\
    \end{longtable}%
    \end{center}%
The split serological types are obtained from \cite{Holdsworth2009}. The left column indicates the supertypes
    defined by the cluster tree. Remark on the labels for the alleles:
    ``(U.)'' stands for ``undefined'' marked by the experts in \cite{Holdsworth2009};
    ``(s.s.)'' indicates that the normal forms of the allele
    is shorter than 81 amino acids; ``($n$)'' with $n=2,3,\cdots$ indicates that
    the normal form is shared by $n$ alleles.

\renewcommand{\arraystretch}{1.1}
\begin{table}[H]
  \centering
\small
    \begin{tabular}{c|c|c}
    \hline\hline
    \multicolumn{3}{c}{\textbf{HLA-DRB1 serological families}}  \\
    \hline
    \textbf{Broad Serotype} & \textbf{Split serotype} & \textbf{Alleles} \\
    \hline
    \multirow{2}{*}{DR1}   & DR1   & DRB1*01 \\\hhline{~--}
          & DR103 & DRB1*0103 \\
          \hline
    \multirow{3}{*}{DR2}   & DR2   & DRB1*1508, *1603 \\\hhline{~--}
          & DR15  & DRB1*15 \\\hhline{~--}
          & DR16  & DRB1*16 \\
          \hline
    \multirow{4}{*}{DR3}   & DR3   & DRB1*0305, *0306, *0307, \\&& *0312, *0314, *0315, *0323 \\\hhline{~--}
          & DR17  & DRB1*0301, *0304, *0310, *0311 \\\hhline{~--}
          & DR18  & DRB1*0302, *0303 \\
          \hline
    DR4   & DR4   & DRB1*04 \\
    \hline
    \multirow{2}{*}{DR5}   & DR11  & DRB1*11 \\\hhline{~--}
          & DR12  & DRB1*12 \\
          \hline
    \multirow{5}{*}{DR6}   & DR6   & DRB1*1416, *1417, *1418 \\\hhline{~--}
          & DR13  & DRB1*13, *1453 \\\hhline{~--}
          & DR14  & DRB1*14, *1354 \\\hhline{~--}
          & DR1403 & DRB1*1403 \\\hhline{~--}
          & DR1404 & DRB1*1404 \\
          \hline
    DR7   & DR7   & DRB1*07 \\
    \hline
    DR8   & DR8   & DRB1*08, *1415 \\
    \hline
    DR9   & DR9   & DRB1*09 \\
    \hline
    DR10  & DR10  & DRB1*10 \\
    \hline\hline
    \multicolumn{3}{c}{\textbf{DRB3/4/5 serological families}}  \\
    \hline
    \multicolumn{2}{c|}{\textbf{Serotype}} & \textbf{Alleles} \\
    \hline
    \multicolumn{2}{c|}{DR51} & DRB5*01,02 \\
    \hline
    \multicolumn{2}{c|}{DR52} & DRB3*01,02,03 \\
    \hline
    \multicolumn{2}{c|}{DR53} & DRB4*01 \\
    \hline\hline
    \end{tabular}%
    \caption{Overview of the broad serological types in connection with the split serological types assigned by WHO. The serological type information listed in this table was extracted from the Tables 4 and 5 given in \cite{Holdsworth2009}. This table summarizes the allele and serotype information given in the first and third columns of Tables 4 and 5.}
  \label{tab:addlabel}%
\end{table}
\renewcommand{\arraystretch}{1}

We are far from claiming to have any definitive answers or final 
statements on these questions of peptide binding and serotype clustering.
Many problems here are left unresolved. For example, the serotype clustering result is
more provocative than otherwise and further studies are needed. 
One could look at more automatic choice of the supertypes, or
develop comparative schemes. One could also study problems of 
phylogenetic trees from this point of view as those of H5N1.
Extending the framework to 3D structures of proteins, 
instead of just amino acid chains is suggested. 
We intend to study these questions ourselves and hope that our study will
persuade others to think about these kernels on amino acid chains.

\section*{Acknowledgment}

The authors would like to thank
Shuaicheng Li for pointing out to us that the portions of
DRB alleles that contact with peptides can be obtain from the non-aligned DRB
amino acid sequences by the use of two markers, ``RFL'' and ``TVQ''. We thank
Morten Nielsen for his criticism on over fitting.

We thank Yiming Cheng for his suggestions on the computer code which
were very helpful for speeding up the algorithm for evaluating $K^3$.
He also discussed with us the influence on HLA--peptide binding prediction
of using different representations of the alleles, and of
adjusting the index $\beta$ in the kernel according to the sequence length.
Although the topics are not included in the paper, they have some potential for future work. 

Also, we appreciate Felipe Cucker for reviewing our draft, making many improvements.

\section*{Appendix}
\appendix
\section{The BLOSUM62-2 Matrix}\label{blosumDataAppendix}

We list the whole BLOSUM62-2 matrix in Table \ref{bls}. Table \ref{aminoA} explains the 
amino acids denoted by the capital letters.
\begin{table}[H]
\begin{center}\footnotesize
\begin{tabular}{c|rrrrrrrrrr}\hline\hline
& A&R&N&D&C&Q&E&G&H&I \\ \hline
A & 3.9029&0.6127&0.5883&0.5446&0.8680&0.7568&0.7413&1.0569&0.5694&0.6325 \\
R & 0.6127&6.6656&0.8586&0.5732&0.3089&1.4058&0.9608&0.4500&0.9170&0.3548 \\
N & 0.5883&0.8586&7.0941&1.5539&0.3978&1.0006&0.9113&0.8637&1.2220&0.3279 \\
D & 0.5446&0.5732&1.5539&7.3979&0.3015&0.8971&1.6878&0.6343&0.6786&0.3390 \\
C &  0.8680& 0.3089& 0.3978& 0.3015&19.5766& 0.3658& 0.2859& 0.4204& 0.3550& 0.6535 \\
Q & 0.7568&1.4058&1.0006&0.8971&0.3658&6.2444&1.9017&0.5386&1.1680&0.3829 \\
E & 0.7413&0.9608&0.9113&1.6878&0.2859&1.9017&5.4695&0.4813&0.9600&0.3305 \\
G & 1.0569&0.4500&0.8637&0.6343&0.4204&0.5386&0.4813&6.8763&0.4930&0.2750 \\
H &  0.5694& 0.9170& 1.2220& 0.6786& 0.3550& 1.1680& 0.9600& 0.4930&13.5060& 0.3263 \\
I & 0.6325&0.3548&0.3279&0.3390&0.6535&0.3829&0.3305&0.2750&0.3263&3.9979 \\
L & 0.6019&0.4739&0.3100&0.2866&0.6423&0.4773&0.3729&0.2845&0.3807&1.6944 \\
K & 0.7754&2.0768&0.9398&0.7841&0.3491&1.5543&1.3083&0.5889&0.7789&0.3964 \\
M & 0.7232&0.6226&0.4745&0.3465&0.6114&0.8643&0.5003&0.3955&0.5841&1.4777 \\
F & 0.4649&0.3807&0.3543&0.2990&0.4390&0.3340&0.3307&0.3406&0.6520&0.9458 \\
P & 0.7541&0.4815&0.4999&0.5987&0.3796&0.6413&0.6792&0.4774&0.4729&0.3847 \\
S & 1.4721&0.7672&1.2315&0.9135&0.7384&0.9656&0.9504&0.9036&0.7367&0.4432 \\
T & 0.9844&0.6778&0.9842&0.6948&0.7406&0.7913&0.7414&0.5793&0.5575&0.7798 \\
W & 0.4165&0.3951&0.2778&0.2321&0.4500&0.5094&0.3743&0.4217&0.4441&0.4089 \\
Y & 0.5426&0.5560&0.4860&0.3457&0.4342&0.6111&0.4965&0.3487&1.7979&0.6304 \\
V & 0.9365&0.4201&0.3690&0.3365&0.7558&0.4668&0.4289&0.3370&0.3394&2.4175 \\
\hline \hline
& L&K&M&F&P&S&T&W&Y&V \\ \hline
A & 0.6019&0.7754&0.7232&0.4649&0.7541&1.4721&0.9844&0.4165&0.5426&0.9365 \\
R & 0.4739&2.0768&0.6226&0.3807&0.4815&0.7672&0.6778&0.3951&0.5560&0.4201 \\
N & 0.3100&0.9398&0.4745&0.3543&0.4999&1.2315&0.9842&0.2778&0.4860&0.3690 \\
D & 0.2866&0.7841&0.3465&0.2990&0.5987&0.9135&0.6948&0.2321&0.3457&0.3365 \\
C & 0.6423&0.3491&0.6114&0.4390&0.3796&0.7384&0.7406&0.4500&0.4342&0.7558 \\
Q & 0.4773&1.5543&0.8643&0.3340&0.6413&0.9656&0.7913&0.5094&0.6111&0.4668 \\
E & 0.3729&1.3083&0.5003&0.3307&0.6792&0.9504&0.7414&0.3743&0.4965&0.4289 \\
G & 0.2845&0.5889&0.3955&0.3406&0.4774&0.9036&0.5793&0.4217&0.3487&0.3370 \\
H & 0.3807&0.7789&0.5841&0.6520&0.4729&0.7367&0.5575&0.4441&1.7979&0.3394 \\
I & 1.6944&0.3964&1.4777&0.9458&0.3847&0.4432&0.7798&0.4089&0.6304&2.4175 \\
L & 3.7966&0.4283&1.9943&1.1546&0.3711&0.4289&0.6603&0.5680&0.6921&1.3142 \\
K & 0.4283&4.7643&0.6253&0.3440&0.7038&0.9319&0.7929&0.3589&0.5322&0.4565 \\
M & 1.9943&0.6253&6.4815&1.0044&0.4239&0.5986&0.7938&0.6103&0.7084&1.2689 \\
F & 1.1546&0.3440&1.0044&8.1288&0.2874&0.4400&0.4817&1.3744&2.7694&0.7451 \\
P &  0.3711& 0.7038& 0.4239& 0.2874&12.8375& 0.7555& 0.6889& 0.2818& 0.3635& 0.4431 \\
S & 0.4289&0.9319&0.5986&0.4400&0.7555&3.8428&1.6139&0.3853&0.5575&0.5652 \\
T & 0.6603&0.7929&0.7938&0.4817&0.6889&1.6139&4.8321&0.4309&0.5732&0.9809 \\
W &  0.5680& 0.3589& 0.6103& 1.3744& 0.2818& 0.3853& 0.4309&38.1078& 2.1098& 0.3745 \\
Y & 0.6921&0.5322&0.7084&2.7694&0.3635&0.5575&0.5732&2.1098&9.8322&0.6580 \\
V & 1.3142&0.4565&1.2689&0.7451&0.4431&0.5652&0.9809&0.3745&0.6580&3.6922 \\
\hline\hline
\end{tabular}
\caption{The BLOSUM62-2 matrix.}\label{bls}
\end{center}
\end{table}

\begin{table}[H]
\begin{center}
\begin{tabular}{lc|lc}\hline\hline
A	&	Alanine 	&	L	&	    Leucine 	\\
R	&	    Arginine 	&	K	&	    Lysine 	\\
N	&	    Asparagine 	&	M	&	    Methionine 	\\
D	&	    Aspartic acid &	F	&	    Phenylalanine 	\\
C	&	    Cysteine 	&	P	&	    Proline 	\\
Q	&	    Glutamine 	&	S	&	    Serine 	\\
E	&	    Glutamic acid &	T	&	    Threonine 	\\
G	&	    Glycine 	&	W	&	    Tryptophan 	\\
H	&	    Histidine 	&	Y	&	    Tyrosine 	\\
I	&	    Isoleucine 	&	V	&	    Valine 	\\\hline\hline
\end{tabular}
\caption{The list of the amino acids.}\label{aminoA}
\end{center}
\end{table}

From the Introduction, we see that the matrix $Q$ can be recovered from the BLOSUM62-2 once the marginal probability 
vector $p$ is available. The latter vector is obtained by
\begin{eqnarray*}
p = (\mbox{[BLOSUM62-2]})^{-1} v_1,
\end{eqnarray*}
where $v_1 = (1,\cdots,1)\in\mathbb{R}^{20}$ is a vector with all its coordinate being $1$. 
The matrix $Q$ can be obtained precisely from {\tt http://www.ncbi.nlm.nih.gov/IEB/ToolBox/ 
CPP\_DOC/lxr/source/src/algo/blast/composition\_adjustment/ \\
matrix\_frequency\_data.c\#L391}.

\bibliographystyle{plain}
\bibliography{entropy}

\end{document}